\documentclass[10pt,twocolumn,letterpaper]{article}
\usepackage[pagenumbers]{cvpr} 
\usepackage{url}
\usepackage{booktabs}
\usepackage{multirow}
\usepackage{graphicx}
\usepackage{xspace}

\usepackage{mathtools}
\usepackage{adjustbox}
\usepackage{array}
\usepackage{colortbl}
\usepackage{float,wrapfig}
\usepackage{hhline}
\usepackage{subcaption} 
\usepackage{enumitem}
\usepackage{makecell}
\usepackage{tikz}
\usetikzlibrary{tikzmark}

\usepackage{wrapfig}

\def\tablecite#1{[\citenum{#1}]}
\newcolumntype{g}{>{\columncolor{mitblue}}c}

\newcolumntype{i}{>{\columncolor{gray}}c}

\definecolor{cvprblue}{rgb}{0.21,0.49,0.74}
\definecolor{mitblue}{rgb}{0.88,0.95,0.96}
\definecolor{gold}{rgb}{0.75,0.6,0.12}
\colorlet{shadecolor}{gray!40}

\usepackage{amsmath,amsfonts,bm}

\def\eqref#1{equation~\ref{#1}}

\def\1{\bm{1}}

\DeclareMathAlphabet{\mathsfit}{\encodingdefault}{\sfdefault}{m}{sl}
\SetMathAlphabet{\mathsfit}{bold}{\encodingdefault}{\sfdefault}{bx}{n}

\usepackage{amsmath,amsfonts}
\usepackage{algorithmic}
\usepackage{algorithm}
\usepackage{array}
\usepackage{textcomp}
\usepackage{stfloats}
\usepackage{url}
\usepackage{verbatim}
\usepackage{graphicx}
\usepackage{xspace}
\usepackage{booktabs} 
\usepackage{multirow} 
\usepackage{amsmath}
\usepackage{xcolor}
\usepackage{colortbl}
\usepackage{color}
\usepackage[normalem]{ulem}
\usepackage{tabularray}
\usepackage{longtable}
\usepackage{amssymb}
\usepackage{enumitem}
\usepackage{cuted}

\definecolor{cvprblue}{rgb}{0.21,0.49,0.74}
\usepackage[pagebackref,breaklinks,colorlinks,allcolors=cvprblue]{hyperref}

\title{Dora: Sampling and Benchmarking for 3D Shape Variational Auto-Encoders}

\author {
    Rui Chen\textsuperscript{\rm 1,2}\quad
    Jianfeng Zhang\textsuperscript{\rm 2}$^{\dag}$\quad
    Yixun Liang\textsuperscript{\rm 1,3}\quad
    Guan Luo\textsuperscript{\rm 2,4}\quad
    Weiyu Li\textsuperscript{\rm 1,3}\\
    Jiarui Liu\textsuperscript{\rm 1,3}\quad
    Xiu Li\textsuperscript{\rm 2}\quad
    Xiaoxiao Long\textsuperscript{\rm 1,3}\quad
    Jiashi Feng\textsuperscript{\rm 2}\quad
    Ping Tan\textsuperscript{\rm 1,3}$^{\dag}$\\
    \small $^{\dag}$Corresponding authors \\
    \textsuperscript{\rm 1}The Hong Kong University of Science and Technology\quad
    \textsuperscript{\rm 2}ByteDance Seed\quad \\
    \textsuperscript{\rm 3}LightIllusions\quad
    \textsuperscript{\rm 4}Tsinghua University \\
    \url{https://aruichen.github.io/Dora}
}

\begin{document}
\maketitle
\begin{figure*}
\center
  \includegraphics[width=\textwidth]{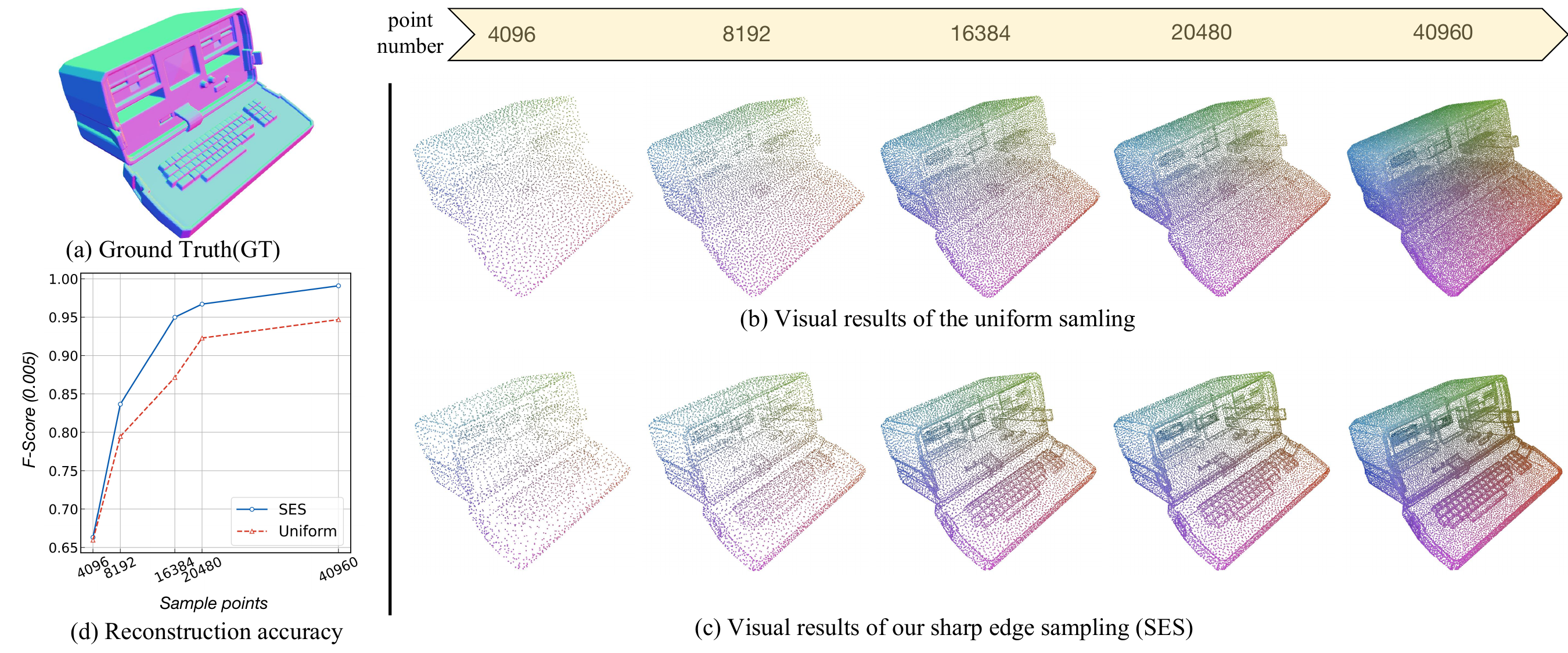}
  \caption{
    Sampling strategy comparison. Given the ground truth mesh shown in (a), we visualize point clouds produced by uniform sampling in (b) and those generated by our proposed Sharp Edge Sampling (SES) in (c), at various sampling rates. In (d), we compare the reconstruction accuracy trained with our SES and the uniform sampling using the F-score metric. The comparison demonstrates SES consistently outperforms uniform sampling under varying sampling rates, as the point clouds generated by SES are more effective in capturing the salient features of the object.
  }
  \label{fig:teaser}
\end{figure*}

\begin{abstract}

Recent 3D content generation pipelines commonly employ Variational Autoencoders (VAEs) to encode shapes into compact latent representations for diffusion-based generation. However, the widely adopted uniform point sampling strategy in Shape VAE training often leads to a significant loss of geometric details,  limiting the quality of shape reconstruction and downstream generation tasks.
We present Dora-VAE, a novel approach that enhances VAE reconstruction through our proposed sharp edge sampling strategy and a dual cross-attention mechanism. By identifying and prioritizing regions with high geometric complexity during training, our method significantly improves the preservation of fine-grained shape features. Such sampling strategy and the dual attention mechanism enable the VAE to focus on crucial geometric details that are typically missed by uniform sampling approaches.
To systematically evaluate VAE reconstruction quality, we additionally propose Dora-bench, a benchmark that quantifies shape complexity through the density of sharp edges, introducing a new metric focused on reconstruction accuracy at these salient geometric features. Extensive experiments on the Dora-bench demonstrate that Dora-VAE achieves comparable reconstruction quality to the state-of-the-art dense XCube-VAE while requiring a latent space at least 8$\times$ smaller (1,280 vs. $>$ 10,000 codes). Project page: \url{https://aruichen.github.io/Dora}.

\end{abstract}    
\section{Introduction}
\label{sec:introduction}

3D content creation is vital to delivering realistic and immersive experiences in various industries, including games, movies, and AR/VR. However, traditional 3D modeling typically demands significant expertise and manual effort, making it time-consuming and challenging, especially for non-expert users. Recent advances in AI-powered 3D content generation methods~\cite{hong2024lrmlargereconstructionmodel,camdm,zhang2023temo,liu2023syncdreamer,chen2022tango,lan2024ln3diff,EnVision2023luciddreamer,liu2023one2345,TripoSR2024,cao2024avatargo,cao2024dreamavatar,wang2022rodingenerativemodelsculpting} have transformed the field, making it more accessible to many more users.

Following the success of text-to-image generation models~\cite{ho2020denoising,rombach2022high,zhang2023adding,chen2023pixartalpha,chen2024pixartdelta}, recent 3D content creation approaches~\cite{wu2024direct3d,zhang2024clay,ren2024xcube} adopt a two-stage pipeline: encoding 3D shapes into a latent space using variational auto-encoders (VAEs), followed by training a latent diffusion model. The performance of such a generative pipeline heavily relies on the VAE's capability to faithfully encode and reconstruct 3D shapes.

Existing 3D VAEs operate by sampling points on mesh surfaces for shape encoding and then reconstructing the original 3D meshes by its decoder. This process faces unique challenges compared to 2D image VAEs where the input image is fully observable. In comparison, the sampled point cloud often cannot capture all the necessary shape information, which could harm the performance of 3D VAEs. 

Volume-based method~\cite{ren2024xcube} leverages sparse convolution \cite{williams2024fvdb} to process millions of voxelized points for high-fidelity reconstruction. Its dense sampling captures precise shape information. However, this method produces large latent codes (commonly $>$ 10,000 tokens), which significantly complicate the training of diffusion models. On the other hand, vector-set (Vecset) methods~\cite{shape2vecset,zhang2024clay,zhao2023michelangelo,li2024craftsman} use transformers to achieve compact latent representations (hundreds to thousands of tokens), enabling efficient diffusion~\cite{zhang2024clay,li2024craftsman}. However, due to the quadratic complexity of transformer networks, it often only samples a few thousand points to represent a 3D shape, which leads to information loss and performance degradation. Therefore, we seek to improve the reconstruction quality of Vecset-based VAEs and maintain their compact representation.

We begin by analyzing the shape reconstruction capability of Vecset-based VAEs.
Through careful analysis, we find these methods have limited reconstruction performance, which stems from their commonly used uniform sampling. When computational constraints limit the total number of sampled points, uniform sampling fails to prioritize geometrically salient regions, leading to the loss of fine details. 
To validate this observation, we experiment on a 3D mesh with intricate geometric details (e.g., keyboard buttons) shown in Figure~\ref{fig:teaser} (a). We visualize the point cloud with different sampling strategies at various sampling densities in (b) and (c). As demonstrated in Figure~\ref{fig:teaser} (b), even with increasing sampling rates, uniform sampling fails to preserve sharp features like keyboard buttons. This simple experiment confirms that uniform sampling fundamentally limits the capturing of geometric details, which in turn affects the VAE's reconstruction capability and the result quality of the learned diffusion models.

Inspired by the success of importance sampling in various geometric processing tasks~\cite{IDIS, wu_2023_attention_edge}, we introduce a similar strategy for 3D VAE training. While existing importance sampling methods focus on down-sampling point clouds, our task requires sampling points directly from mesh surfaces for shape VAE training~\cite{wu2024direct3d,zhang2024clay}. This fundamental difference necessitates a new sampling approach specifically designed for preserving geometric features from mesh representations.

To address this challenge, we propose a \textbf{Sharp Edge Sampling (SES)} algorithm that adaptively samples points based on geometric saliency. Specifically, SES first identifies edges with significant dihedral angles on the mesh, indicating regions of high geometric complexity. It then samples points along these salient regions while maintaining a balance with uniformly sampled points to capture the overall structure. This approach ensures comprehensive coverage of both fine details and global geometry.

Building upon SES, we present Dora-VAE, a novel method achieving high-fidelity reconstruction while maintaining compact latent representations. To fully leverage these detail-rich point clouds sampled by our SES, we design a dual cross-attention architecture that effectively processes both salient and uniform regions during encoding.  As shown in Figure~\ref{fig:teaser}, our method significantly outperforms uniform sampling in preserving shape details on the keyboard (c), with consistent improvements in F-Score across different sampling rates (d). 

The common evaluation protocol for 3D VAEs is also biased. It typically uses a set of randomly selected 3D shapes, and employs general metrics (e.g., F-score, Chamfer distance) to measure the shape reconstruction quality. However, we argue it is necessary to divide the test shapes into sub-classes of different shape complexity to better evaluate these 3D VAEs. To facilitate the evaluation of 3D VAEs, we further introduce Dora-Bench, a new benchmark based on existing datasets with our novel Sharp Normal Error (SNE) metric. Dora-Bench categorizes test shapes based on their geometric complexity and SNE focuses on measuring the reconstruction quality of salient geometric features. This combination enables a more rigorous assessment of 3D VAEs.

Extensive experiments demonstrate that our Dora-VAE achieves superior results. When integrated into downstream 3D diffusion models, it significantly enhances the quality of generated 3D shapes, which validates that our novel sampling strategy and dual cross-attention architecture effectively preserve geometric details with compact latent spaces. To summarize, our main contributions include:
1) We propose Dora-VAE, a novel 3D VAE model for high-quality reconstruction with compact latent representations, accompanied by Dora-Bench, a comprehensive benchmark for evaluating 3D VAEs.
2) We introduce, for the first time, importance sampling to the task of 3D VAE learning and propose a Sharp Edge Sampling (SES) algorithm to prioritize geometrically salient regions. Building on SES, we design a dual cross-attention architecture to effectively encode these detail-rich point clouds.
3) We develop a systematic evaluation benchmark based on existing datasets with our novel Sharp Normal Error (SNE) metric 
that specifically assesses reconstruction accuracy of fine geometric details, enabling more rigorous evaluation than conventional random sampling approaches.

\section{Related work}
\textbf{Importance Sampling in Point Clouds.} 
Importance sampling techniques have been widely used in point cloud processing tasks~\cite{IDIS,wu_2023_attention_edge}. 
For instance, APES~\cite{wu_2023_attention_edge} proposes attention-based sampling for point cloud classification and segmentation. 
However, these methods operate directly on point clouds rather than meshes, making them less suitable for VAE-based shape representation where preserving complete geometric information is crucial.

\noindent\textbf{3D Shape VAEs.} 
Recent 3D shape VAEs follow two main approaches: volume-based and vector set-based. Volume-based methods~\cite{xiong2024octfusion, ren2024xcube} like XCube~\cite{ren2024xcube} use sparse convolution to encode voxelized surfaces, achieving high reconstruction quality but requiring large latent codes ($>$ 10,000 tokens). While these methods excel at preserving geometric details, their large latent spaces pose significant challenges for downstream diffusion model training.
In contrast, vector set-based approaches~\cite{shape2vecset, wu2024direct3d, li2024craftsman, zhang2024clay,zhao2023michelangelo} encode uniformly sampled surface points using transformers, producing highly compact latent spaces that are particularly suitable for diffusion models. However, these methods often struggle with geometric detail preservation, especially in regions with complex surface features. Our analysis reveals that this limitation primarily stems from their uniform sampling strategy: when computational constraints restrict the total number of processable points, uniform sampling fails to prioritize geometrically significant regions, leading to insufficient capture of fine details. This information loss at the sampling stage fundamentally limits these methods' ability to learn and preserve intricate geometric features.

\noindent\textbf{3D Content Creation.} 
Current 3D generation methods can be categorized into three groups. 
Optimization-based methods~\cite{poole2022dreamfusion,lin2023magic3d, fantasia3d, wang2023prolificdreamer,RichDreamer, sweetdreamer, DreamGaussian, shi2023MVDream,liu2023syncdreamer, liu2023sherpa3d}, pioneered by DreamFusion~\cite{poole2022dreamfusion} utilize score distillation sampling (SDS) to optimize 3D representations~\cite{mildenhall2020nerf,kerbl20233d, shen2021deep} using 2D diffusion model priors. While these methods can achieve photorealistic results, they suffer from slow generation speed, training instability, and often struggle to maintain geometric consistency.
Large reconstruction models, like LRM~\cite{LRM} and follow-up works~\cite{xu2024instantmesh,li2023instant3d, wang2024crm, tang2024lgm,liu2024meshformer,wang2023pf} employ large-scale sparse-view reconstruction for efficient 3D generation. However, their lack of explicit geometric priors often leads to compromised geometric fidelity and inconsistent surface details.
3D native generative models \cite{shape2vecset, li2024craftsman, wu2024direct3d, zhang2024clay,zhao2023michelangelo}, represented by 3DShape2VecSet~\cite{shape2vecset}, adopt a two-stage approach: first training a 3D VAE to encode shapes into latent space, then training a conditional latent diffusion model for generation. 
This approach ensures better geometric consistency through the VAE's built-in geometric constraints. However, the quality of generated shapes is fundamentally limited by the VAE's reconstruction capability. Recent works~\cite{li2024craftsman,zhang2024lagem} have shown that improving VAE reconstruction directly enhances downstream generation quality, which motivates our focus on advancing VAE design.

\section{Method}
\label{sec:Method}

In this section, we present \textbf{Dora-VAE} for high-quality 3D reconstruction, and \textbf{Dora-Bench} for 3D VAE evaluation.
We first briefly review 3DShape2VecSet \cite{shape2vecset} in Section \ref{sec:preliminary}, the foundation of our method and then detail our key innovations in Section \ref{sec:dora_vae} and Section \ref{sec:dora_bench}.

\subsection{Preliminary: 3DShape2VecSet}
\label{sec:preliminary}
3DShape2VecSet \cite{shape2vecset} introduces a transformer-based 3D VAE that encodes uniformly sampled surface points into compact latent codes. 
Given a 3D surface S, their pipeline consists of three key steps:
\begin{itemize}
    \item Surface Sampling: Uniformly sample $N_d$ points on the surface S using Poisson disk sampling \cite{PoissionSampling} to obtain a dense point cloud $P_d$, then downsample it to $N_{s}$ points via Farthest Point Sampling (FPS) \cite{moenning2003fast} to get a sparse point cloud $P_{s}$:
        \begin{align}
        \label{eq:qkv}
            P_d = \{p_d^i \in S \mid i = 1,...,N_d\}, P_{s} = \text{FPS}(P_d, N_{s}).
        \end{align}

    \item Feature Encoding: Compute the point cloud feature $C$ via the cross-attention between $P_{s}$ and $P_d$, followed by some self-attention layers to generate the latent code $z$:
    \begin{equation}
         C=\text{CrossAttn}(P_{s}, P_d, P_d), z = \text{SelfAttn}(C).
    \end{equation}

    \item Geometry Decoding: Further decode $z$ through self-attention layers and predict occupancy values using randomly sampled spatial query points \( Q_{space} \in \mathbb{R}^3 \):
    \begin{equation}
    \label{eq:occupancy}
        \hat{O} = \text{CrossAttn}(Q_{space}, \text{SelfAttn}(z)).
    \end{equation}
\end{itemize}
While this method generates compact latent codes for diffusion, the uniform sampling limits its ability to capture fine geometric details. Our work addresses this limitation by carefully designed sampling and encoding strategies.

\begin{figure}[t!]
  \centering
  \includegraphics[width=\linewidth]{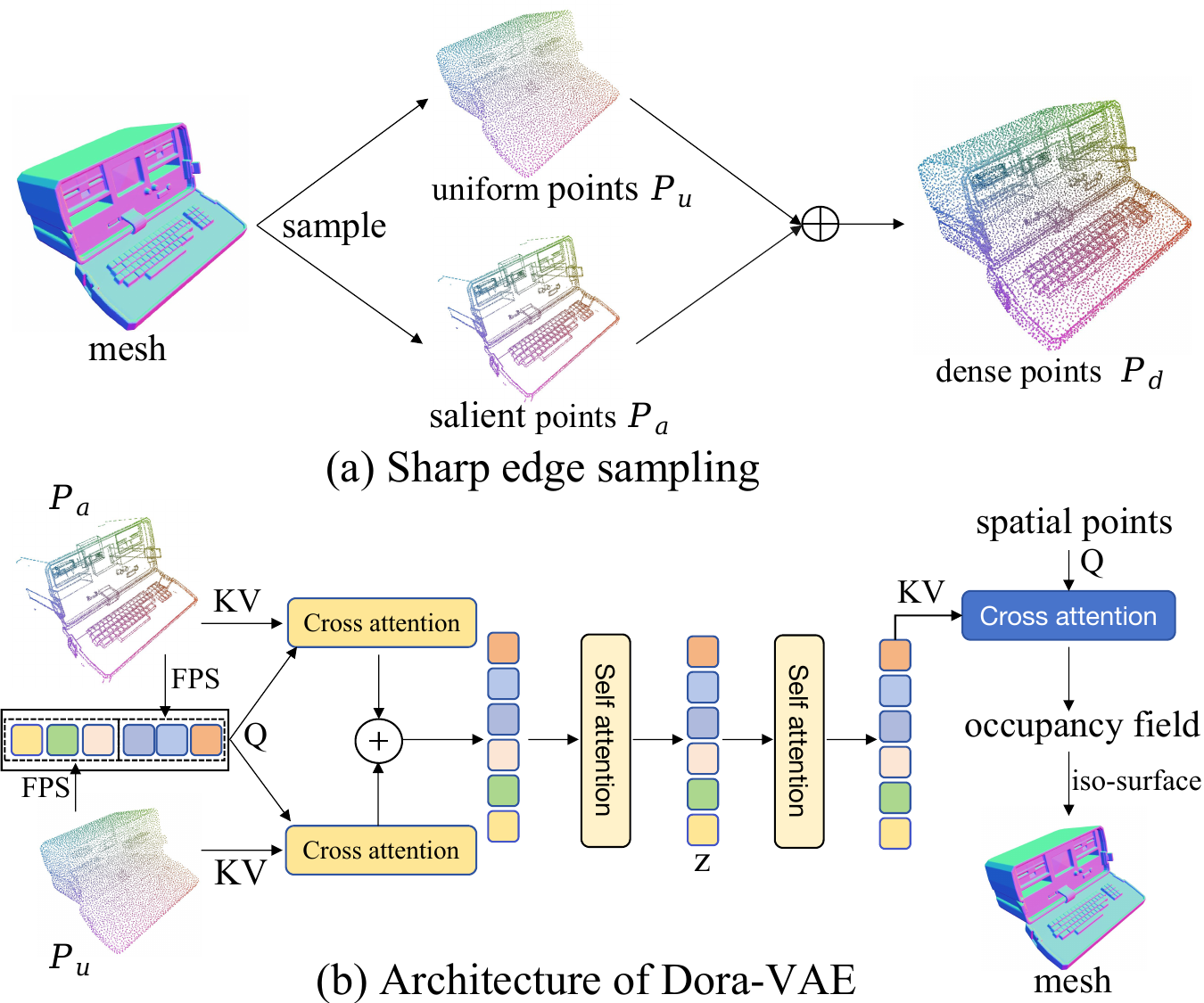}
  \caption{
    \textbf{Overview of Dora-VAE.} (a) We utilize the proposed sharp edge sampling technique to extract both salient and uniform points from the input mesh. These points are then combined with dense points, effectively capturing both salient regions and smooth areas. (b) To enhance the encoding of point clouds sampled through sharp edge sampling, we design a dual cross-attention architecture.
  }
  \vspace{-4mm}\label{fig:3dnativegeneration}
\end{figure}

\subsection{Dora-VAE}
\label{sec:dora_vae}
Figure \ref{fig:3dnativegeneration} gives an overview of our pipeline. For each input mesh, we augment the uniformly sampled point cloud $P_u$ with more important points $P_a$ sampled by our proposed sharp edge sampling strategy, which forms the dense point cloud $P_d$. During the encoding process, we compute the attention for $P_u$ and $P_a$ separately via a simple-yet-effective dual cross attention mechanism and sum the results for self-attention to compute the latent code $z$. Our VAE training largely follows that of 3DShape2VecSet~\cite{shape2vecset}, which is supervised by a loss evaluated on the occupancy field. 

\subsubsection{Sharp Edge Sampling (SES)}
\label{sec:importance_sampling}
We propose SES to effectively sample points from geometrically salient regions. To ensure surface coverage, we also sample points uniformly. Our final sampled point cloud $P_d$ combines uniformly sampled points $P_u$ with points specifically sampled from salient regions $P_a$ as $P_d = P_u \cup P_a$. 
Our method computes salient points $P_a$ through two steps: detecting salient edges and sampling points from these regions.

\noindent\textbf{Salient Edges Detection.}
Given a triangular mesh, we identify a set of salient edges $\Gamma$ by analyzing dihedral angles between adjacent faces, which calculates the angle between the normal vectors of adjacent faces, providing a direct measure of surface curvature at mesh edges. For each edge $e$ shared by adjacent faces $f_1$ and $f_2$, we compute the dihedral angle $\theta_e$ as: 
\begin{equation}
    \theta_e = \arccos \left( \frac{\mathbf{n}_{f_1} \cdot \mathbf{n}_{f_2}}{\|\mathbf{n}_{f_1}\| \|\mathbf{n}_{f_2}\|} \right),
    \label{eq:dihedral_angle}
\end{equation}
where \( \mathbf{n}_{f_1} \) and \( \mathbf{n}_{f_2} \) are the normals of $f_1$ and $f_2$.
The salient edge set $\Gamma$ contains all edges with a dihedral angle exceeding a predefined threshold \( \tau \):

\begin{equation}
    \Gamma = \{ e \mid \theta_e > \tau \}
\end{equation}
Let \( N_{\Gamma} = | \Gamma | \) represent the number of the salient edges.

\noindent\textbf{Salient Points Sampling.}
For each salient edge $e \in \Gamma$, we collect its two vertices $v_{e,1}$ and $v_{e,2}$ into a salient vertex set $P_{\Gamma}$:
\begin{equation}
P_{\Gamma} = \{ v_{e,1}, v_{e,2} \mid e \in \Gamma \},
\end{equation}
where duplicate vertices from connecting edges are included only once. Let $N_V = |P_{\Gamma}|$ denote the number of unique vertices in $P_{\Gamma}$.

Given a target number of salient points \( N_{\text{desired}} \), we generate the salient point set \( P_a \) based on the available salient vertices: 

\begin{equation}
    P_a =
    \begin{cases}
        \text{FPS}(P_{\Gamma}, N_{\text{desired}}), & \text{if } N_{\text{desired}} <= N_V, \\
        P_{\Gamma} \cup P_{\text{interpolated}}, & \text{if } 0 < N_V  < N_{\text{desired}}, \\
        \emptyset, & \text{if } N_V = 0.
    \end{cases}
\end{equation}
When we have excess salient vertices ($N_{\text{desired}} \leq N_V$), we use FPS to downsample $P_{\Gamma}$ to obtain $P_a$. For cases with insufficient salient vertices (\( N_{V} < N_{\text{desired}} \)), we include all vertices from $P_{\Gamma}$ and supplement with additional points $P_{\text{interpolated}}$. These additional points are generated by uniformly sampling $ (N_{\text{desired}}-N_V)/N_{\Gamma}$ points along each salient edge in $\Gamma$, ensuring comprehensive coverage of salient features. When no salient edges are detected ($N_V = 0$), $P_a$ remains empty.

\begin{figure}
    \centering
    \includegraphics[width=\linewidth]{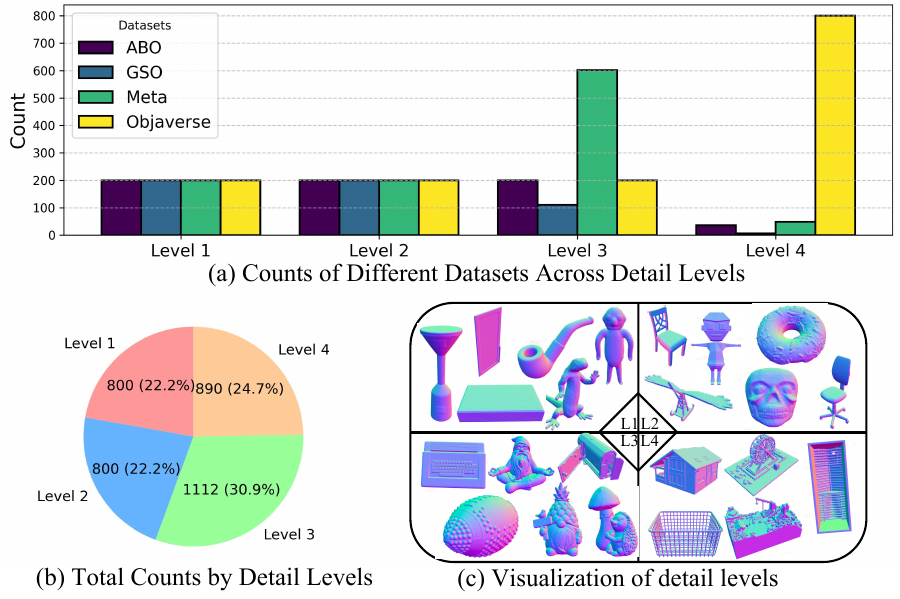}
    \caption{Our proposed benchmark include 3D shapes from the ABO~\cite{collins2022abo}, GSO~\cite{downs2022google}, Meta~\cite{meta_dtc}, and Objaverse~\cite{objaverse} datasets. (a) The histogram of different datasets across different shape complexities. (b) The pie chart of the total counts by shape complexities. (c) Sample shapes of different shape complexities.}
    \label{fig:difficulty_levels}
\end{figure}

\subsubsection{Dual Cross Attention}
\label{subsec:dual_cross_attention}
Given the point clouds \( P_d \) produced by our SES strategy, we design a dual cross-attention architecture to effectively encode both uniform and salient regions.  
Following 3DShape2VecSet~\cite{shape2vecset}, we first downsample  $P_{u}$ and $P_{a}$ seperately using FPS:
\begin{equation}
P_s = \text{FPS}(P_u, N_{s,1}) \cup \text{FPS}(P_a, N_{s,2}),
\end{equation}
where $N_{s,1}$ and $N_{s,2}$ is the  number of downsampled point clouds from  $P_{u}$ and $P_{a}$, respectively.
We then compute cross-attention features separately for uniform and salient points as follows:
\begin{align}
C_u &= \text{CrossAttn}(P_s, P_u, P_u) \\
C_a &= \text{CrossAttn}(P_s, P_a, P_a)
\end{align}
The final point cloud feature  \( C \) combines both attention results:
\begin{equation}
   C = C_{u} + C_{a}. 
\end{equation}
This dual attention design enables separate focus on uniform and salient regions during feature extraction. 
Following 3DShape2VecSet~\cite{shape2vecset}, we use $C$ to predict the occupancy field  \( \hat{O} \) through self-attention blocks. The whole model with parameters $\psi $ are optimized using MSE loss:
\begin{equation}\label{EqnSDS}
    \nabla_\psi  \mathcal{L}_{\text{MSE}}(\hat{O},O) = \mathbb{E} \left [2(\hat{O} - O)\frac{\partial \hat{O}}{\partial \psi} \right ].
\end{equation}

\begin{figure}
    \centering
    \includegraphics[width=\linewidth]{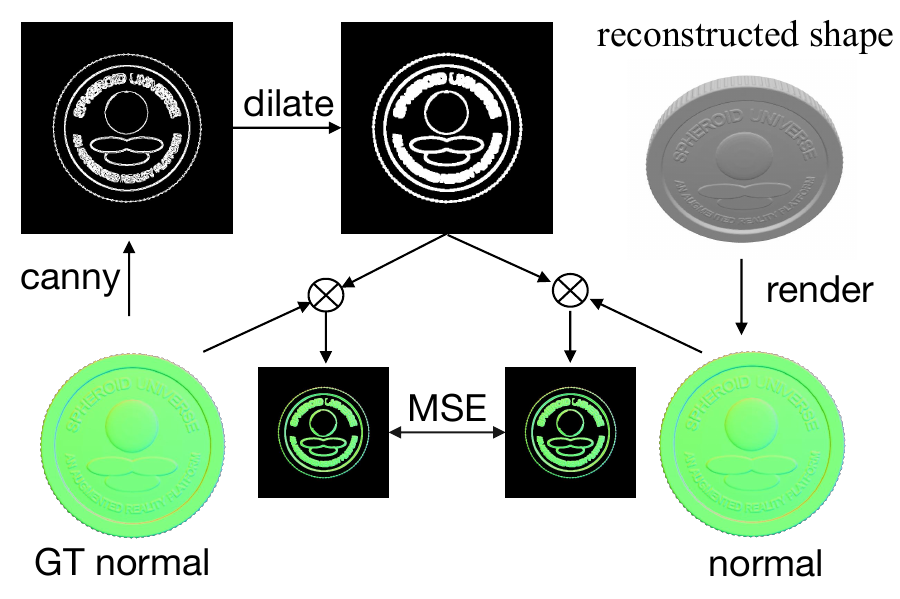}
    \caption{The process of computing sharp normal errors (SNE). We compute MSE loss in the sharp regions of the normal.}
    \label{fig:SNE}
\end{figure}

\subsection{Dora-Bench}
\label{sec:dora_bench}

\begin{figure*}
\center
  \includegraphics[width=\textwidth]{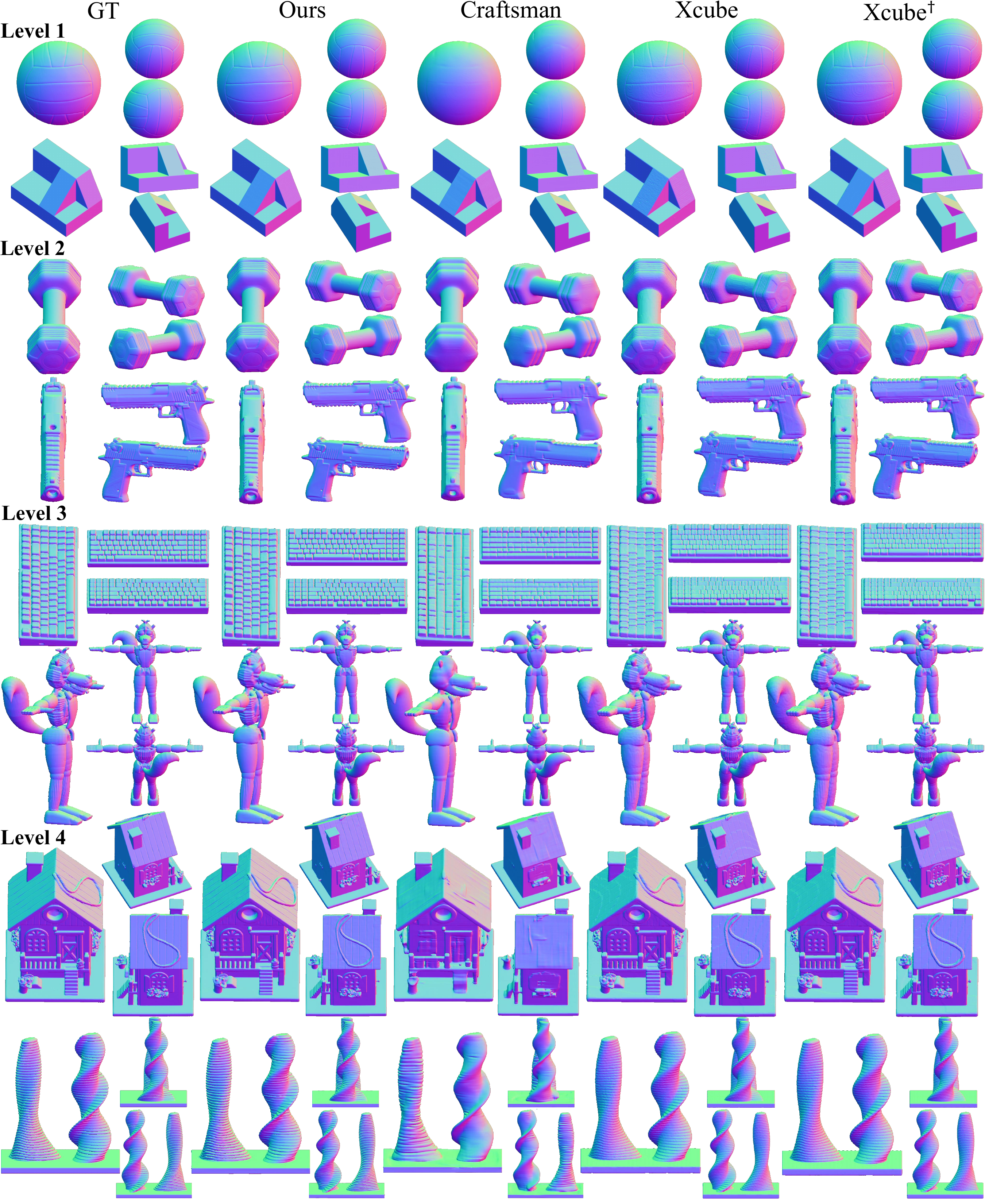}
  \caption{
    Qualitative comparison of the VAE reconstruction results. $^\dag$ indicates the fine-tuning model that uses the same training data as ours.
  }
  \label{fig:qualitative_comparison}
\end{figure*}

\subsubsection{Geometric Complexity-based Evaluation}
\label{sec:evaluation_benchmark}

To enable a more rigorous evaluation of VAE performance, we propose Dora-bench, a benchmark that systematically categorizes test shapes based on their geometric complexity. Unlike previous methods that use randomly selected test sets, we measure shape complexity using the number of salient edges $N_{\Gamma}$ (Section \ref{sec:importance_sampling}) and classify shapes into four levels:
\begin{itemize}
    \item \textbf{Level 1 (Less Detail)}: \( 0 < N_{\Gamma} \leq 5000 \);
    \item \textbf{Level 2 (Moderate Detail)}: \( 5000 < N_{\Gamma} \leq 10000 \);
    \item \textbf{Level 3 (Rich Detail)}: \( 10000 < N_{\Gamma} \leq 50000 \);
    \item \textbf{Level 4 (Very Rich Detail)}: \( N_{\Gamma} > 50000 \).
\end{itemize}

We curate test shapes from multiple public datasets including GSO~\cite{downs2022google},  ABO~\cite{collins2022abo},  Meta~\cite{meta_dtc}, and  Objaverse~\cite{objaverse} to ensure diverse geometric complexities. Figure~\ref{fig:difficulty_levels} shows the distribution of shapes across complexity levels (a,b) and example meshes from each level (c). Please refer to our supplementary materials for more examples.

\subsubsection{Sharp Normal Error (SNE)}
\label{sec:sne}

Building on our Dora-bench, we further introduce Sharp Normal Error (SNE) to evaluate reconstruction quality in salient regions. While existing metrics like Chamfer Distance and F-Score capture overall shape similarity, they fail to specifically assess the preservation of fine geometric details. SNE addresses this limitation by measuring normal map differences between reconstructed and ground truth shapes in geometrically significant areas. 
As illustrated in Figure \ref{fig:SNE}, we render normal maps of the ground truth shape from multiple viewpoints and identify salient regions using Canny edge detection. These regions are dilated to create evaluation masks. The final SNE metric is computed as the Mean Squared Error between ground truth and reconstructed normal maps within the masked areas. This process enables focused evaluation of how well VAEs preserve sharp geometric features during reconstruction.

\begin{table*}[t]
\footnotesize\centering\setlength{\tabcolsep}{2pt}
\begin{tabular}{l | g | g g g g | g g g g | g g g g| g g g g}
\toprule
\rowcolor{white} & & \multicolumn{4}{c |}{$\uparrow$ F-score(0.01) $\times$ 100}& \multicolumn{4}{c |}{$\uparrow$ F-score(0.005) $\times$ 100}  & \multicolumn{4}{c |}{$\downarrow$ CD $\times$ 10000}  &\multicolumn{4}{c}{$\downarrow$ SNE $\times$ 100} \\

\rowcolor{white} \multirow{-2}{*}{Methods} & \multirow{-2}{*}{LCL} & L1 & L2 & L3 & L4 & L1 & L2 & L3 & L4 &L1 & L2 & L3 & L4 &L1 & L2 & L3 & L4 \\
\midrule
\multirow{-1}{*}{Xcube~\tablecite{ren2024xcube}} &  $>$10000  & 98.968 &   98.799 &  98.615 & 98.226 & 95.525 & 93.872 & 92.322 & 85.365 & 6.315 & 6.288 & 7.935 & 9.926 & 1.579 & 1.432 & 1.430 & 1.679\\
\midrule
\rowcolor{white} \multirow{-1}{*}{Xcube$^\dag$~\tablecite{ren2024xcube}} & $>$10000   & 99.393 &  99.794 &  99.824 & 99.079 & 96.753 &  95.535 & 93.422 & 87.365 & 4.015 & 4.142 & 5.740 & 7.627 & 1.543 & 1.408 & 1.259 & 1.639\\

\midrule
\midrule
\multirow{-1}{*}{Craftsman~\tablecite{li2024craftsman}} &  256  & 98.016 & 95.874 & 91.756 & 81.739  & 87.994 & 82.549 & 73.000 & 57.379 & 4.389 & 9.129 & 14.530 & 33.441 & 1.906 & 1.873 & 2.191 & 3.933\\
\midrule
\rowcolor{white} &  1280  & 99.964 & 99.925 & 99.678 & 97.890 & 96.561 & 95.975 & 91.618 & 83.124 & 2.236 & 2.506 & 4.444 & 6.432 & 1.448 & 1.215 & 1.205 & 1.828\\
\multirow{-2}{*}{\textbf{Ours}} \multirow{-2}{*}{\makecell{\tiny{w/o DCA}\\ \tiny{w/o SES,DCA}}} & 1280   & 99.944 &  99.814 & 97.294 & 96.779 & 95.977 &  94.623 & 88.406 & 79.240 & 2.422 & 2.983 & 3.980 & 6.196 & 1.496 & 1.313 & 1.352 & 2.207\\   
\midrule
\rowcolor{white} &  256  & 99.507 & 98.986 & 96.669 & 89.577 & 93.272 & 90.466 & 82.386 & 68.669 & 3.356 & 5.202 & 10.276 & 24.527 & 1.555 & 1.410 & 1.618 & 3.035 \\
\multirow{-2}{*}{\textbf{Ours}}\multirow{-2}{*}{~~~\makecell{\tiny{full}}} & 1280   & \textbf{99.988} &  \textbf{99.955} &  \textbf{99.880} & \textbf{99.170} & \textbf{97.038} &  \textbf{96.831} & \textbf{93.458} & \textbf{87.473} & \textbf{2.097} & \textbf{2.500} & \textbf{3.945} & \textbf{5.265} & \textbf{1.433} & \textbf{1.186} & \textbf{1.137} & \textbf{1.579} \\

\bottomrule
\end{tabular}

\begin{tikzpicture}[overlay, remember picture, shorten >=.5pt, shorten <=.5pt, transform canvas={yshift=.25\baselineskip}]

\end{tikzpicture}
\vspace{-5pt}
\caption{Quantitative comparison in Dora-bench. $^\dag$ indicates the fine-tuning model that uses the same training data as ours.}
\label{tab:quantitative_comparison}
\end{table*}

\section{Experiments}
We conducted intensive experiments to validate the effectiveness of our proposed 3D VAE and compare it with other state-of-the-art methods.

\subsection{Implementation Details}
\label{sec:implementation}
We follow CLAY~\cite{zhang2024clay} for mesh preprocessing to ensure watertight 3D models. Our VAE is trained on a subset filtered from Objaverse~\cite{objaverse}, containing approximately 400,000 3D meshes. We filter out low-quality meshes with missing faces or severe self-intersections to ensure training stability. Our training is conducted on 32 A100 GPUs for two days using a batch size of 2048 and a learning rate of 5e-5. We employ Flash-Attention-v2 ~\cite{FlashAttentionv2}, mixed-precision training with FP16 and gradient checkpointing~\cite{GradientCheckpoint} to optimize memory usage and training efficiency. For the parameters in sharp edge sampling, we set \( N_{\text{desired}} = 16384 \) and $\tau=30$. We set the low threshold to 20 and the high threshold to 200 for the canny edge detection.

\subsection{Evaluation Setting}
\label{sec:evaluation_setting}
\noindent\textbf{Metrics.} 
We evaluate reconstruction quality by comparing input meshes with their decoded counterparts from different 3D VAEs using 1M sampled points under three metrics:
1) F-score (\( r \)) \cite{fscore}, which reconstruction accuracy by computing precision and recall of point correspondences within distance threshold $r$. 
Specifically, we report F-score (0.01) and F-score (0.005) with shapes normalized $[-1,1]$. 2) Chamfer Distance (CD), which computes the average distance between each reconstructed point and its nearest ground truth point.
3) Sharp Normal Errors (SNE) as proposed in Section~\ref{sec:sne}, which evaluate normal map differences in salient regions.
For fair comparison, we also report Latent Code Length (LCL) as longer codes typically enable better reconstruction. 

\noindent\textbf{Baselines.} 
We compare Dora-VAE with state-of-the-art approaches, including: 1) XCube-VAE~\cite{ren2024xcube}, a volumetric method with larger latent codes; 2) XCube-VAE$^\dag$~\cite{ren2024xcube}, our fine-tuned version of the original XCube-VAE on the same dataset; 3) Craftsman-VAE\cite{li2024craftsman}, which fine-tune the 3DShape2VecSet~\cite{shape2vecset} with shorter latent codes on Objaverse. 
We exclude VAE models from Direct3D~\cite{wu2024direct3d} and CLAY\cite{zhang2024clay} as 
as their implementations were not publicly available at submission time.

\subsection{Qualitative Comparison}
\label{sec:qualitative_comparison}
Figure \ref{fig:qualitative_comparison} shows visual comparisons of different methods across different complexity levels from our Dora-bench dataset. We visualize both ground truth and reconstructed meshes using surface normal coloring to highlight geometric details.
For shapes with lower complexity (L1 and L2), all methods achieve comparable reconstruction quality. However, when dealing with shapes of higher complexity (L3 and L4), the advantages of our method become obvious. 
While XCube-VAE achieves similar visual quality to ours, it requires a significantly larger latent space - more than 8$\times$ the size of ours ($>$ 10,000 dimensions vs. 1,280). 
This substantial reduction in latent code length, while maintaining high reconstruction fidelity, makes our method particularly suitable for training 3D diffusion models. 
In contrast, Craftsman-VAE shows a noticeable degradation in reconstruction quality for complex shapes, failing to capture fine geometric details.
Additional visual comparisons are provided in the appendix.

\subsection{Quantitative Comparison}
\label{sec:quantitative_comparison}
Table \ref{tab:quantitative_comparison} presents quantitative results of different methods across different complexity levels of Dora-bench.  Our method consistently outperforms baselines across all levels, with larger margins on more complex shapes (L3 and L4). 
The advantage is particularly evident in CD metrics, where our method with only 256 latent codes surpasses even our fine-tuned version of XCube-VAE (3.356 vs. 4.015). When using 1280 latent codes, our method further decreases CD to 2.097, achieving a 47.77\% improvement over XCube-VAE$^\dag$.
We attribute XCube-VAE's lower performance to its use of NKSR~\cite{huang2023neural} for mesh extraction, which introduces additional quantization errors.

Notably, our method demonstrates superior performance in preserving geometric details, as reflected by the SNE metric. For example, in L4 shapes where geometric complexity is highest, our method achieves an SNE of 1.579 compared to 1.639 from XCube-VAE$^\dag$, representing a 3.7\% improvement. 
This significant gain in SNE aligns with our qualitative observations in Figure \ref{fig:qualitative_comparison}, where our method better preserves fine details such as sharp edges and complex surface variations, demonstrating the effectiveness of our sharp edge sampling strategy.

\begin{figure}
\center
  \includegraphics[width=0.9\linewidth]{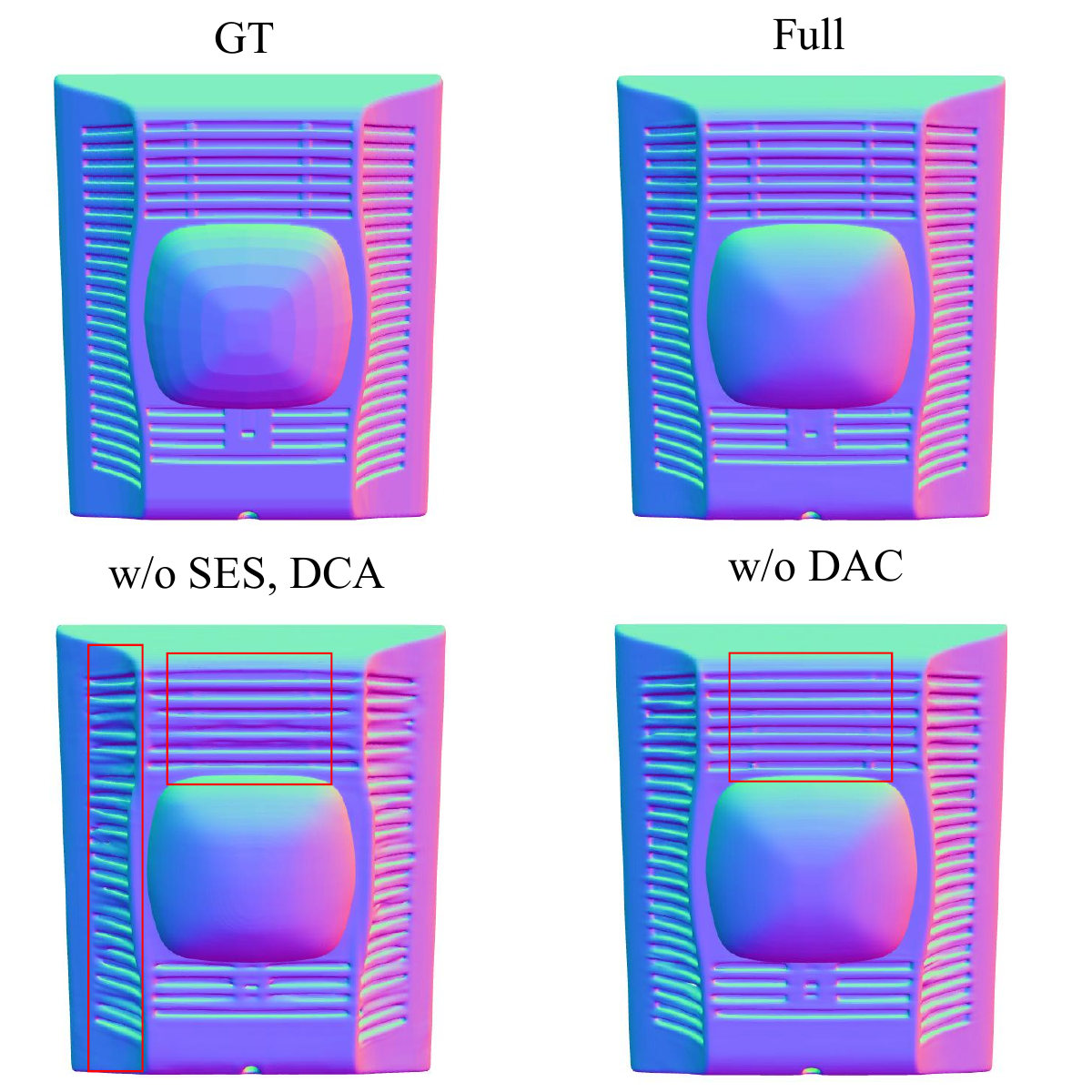}
  \caption{
    Ablation studies of our method. Given the ground truth of mesh,  we employ both our full model and its variations to reconstruct the ground truth mesh, highlighting significant reconstruction discrepancies with {red boxes}.
  }
  \label{fig:ablation}
\end{figure}

\subsection{Ablation Studies}
\label{sec:ablation}
To evaluate the contribution of each component, we compare our full model with two variants under the same training conditions:
\begin{itemize}
    \item Ours w/o SES, DCA. This variant removes both sharp edge sampling (SES) and dual cross attention (DCA), i.e. using only uniformly sampled point clouds with Poisson disk sampling~\cite{PoissionSampling}, while maintaining an equal $N_d$.
    \item Ours w/o DCA. This variant retains SES but removes DCA, i.e. using a single cross attention adopted by~\cite{shape2vecset}.
\end{itemize}
As shown in Figure \ref{fig:ablation} and Table \ref{tab:quantitative_comparison}, our full model consistently outperforms these variants, validating the effectiveness of both components.
\section{Application: Single Image to 3D }
\begin{figure}
\center
  \includegraphics[width=\linewidth]{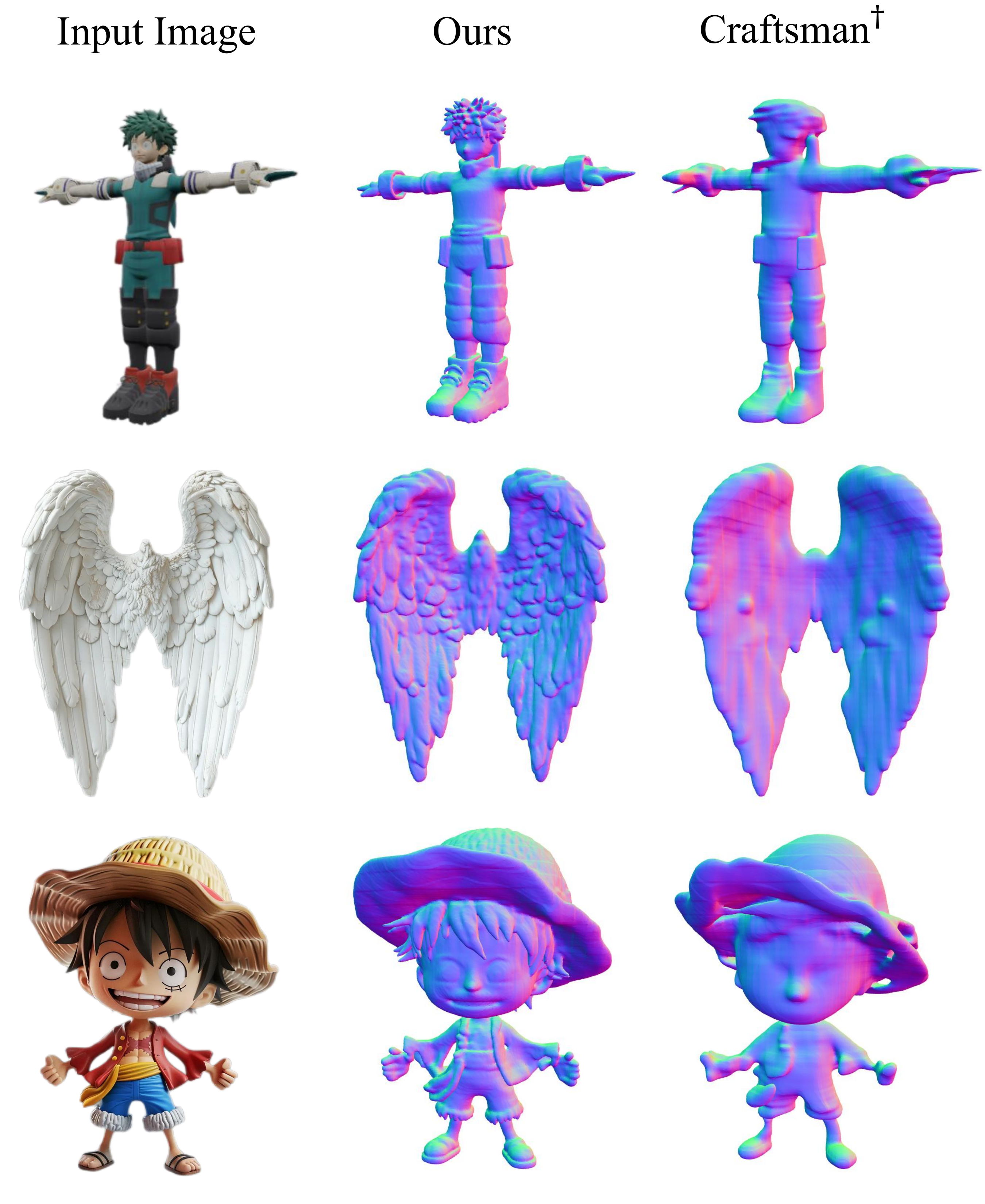}
  \caption{
    The diffusion results of the single image to 3D generation trained on our Dora-VAE and Craftsman$^\dag$. The 3D geometry generated by the diffusion model trained on our proposed Dora-VAE has more details under the same experimental environment.
  }
  \label{fig:diffusion}
  \vspace{-0.6cm}
\end{figure}

We demonstrate the effectiveness of our VAE by applying it to single-image 3D generation through diffusion models. 
Following CLAY~\cite{zhang2024clay}, we implement a latent diffusion model based on the DiT~\cite{peebles2023scalable} architecture.
For a fair comparison, we fine-tune Craftsman-VAE on our dataset (denoted as Craftsman-VAE$^\dag$) since both Craftsman-VAE and 3DShape2VecSet were originally trained on smaller datasets. 
Note that XCube-VAE is excluded from comparison due to its 10,000-dimensional latent codes being impractical for diffusion model training.
Figure~\ref{fig:diffusion} shows some generation results from diffusion models trained with our Dora-VAE and Craftsman-VAE$^\dag$. Both models share identical architecture (0.39B parameters) and training conditions (same dataset, 32 A100 GPUs, 3 days). 
Our Dora-VAE demonstrates significantly better preservation of geometric details in the generated shapes, validating its effectiveness as a foundation for 3D generation tasks.
While further improvements could be achieved with more extensive training data and computational resources, we focus on validating the VAE's capabilities in this work and leave such extensions for future work.

\section{Conclusion}
In this work, we introduce Dora-VAE, a novel VAE designed for high-quality 3D shape compression and reconstruction.
At its core, Dora-VAE introduces sharp edge sampling to effectively capture salient geometric features, complemented by a dual cross-attention architecture that enhances the encoding of these detail-rich point clouds.
To enable more rigorous evaluation of VAE performance, we develop Dora-bench, which systematically categorizes shapes based on geometric complexity and introduces the Sharp Normal Error (SNE) metric for specifically assessing the preservation of fine geometric details. 
Our comprehensive experiments demonstrate that Dora-VAE significantly outperforms existing methods across varying levels of shape complexity.
Furthermore, we show that the improved reconstruction capability of Dora-VAE directly enhances the quality of downstream tasks by applying it to single-image 3D generation. The superior performance in generating geometric details validates our approach of focusing on salient region sampling and encoding for 3D VAE design.

\noindent\textbf{Acknowledgements.} This work is partially supported by the project L0751 between Bytedance and HKUST.

\renewcommand{\thetable}{S\arabic{table}}
\renewcommand{\thefigure}{S\arabic{figure}}
\maketitlesupplementary
\begin{table*}[t]
\footnotesize\centering\setlength{\tabcolsep}{2pt}
\begin{tabular}{l | g | g g g g | g g g g | g g g g| g g g g}
\toprule
\rowcolor{white} & & \multicolumn{4}{c |}{$\uparrow$ F-score(0.01) $\times$ 100}& \multicolumn{4}{c |}{$\uparrow$ F-score(0.005) $\times$ 100}  & \multicolumn{4}{c |}{$\downarrow$ CD $\times$ 10000}  &\multicolumn{4}{c}{$\downarrow$ SNE $\times$ 100} \\

\rowcolor{white} \multirow{-2}{*}{Methods} & \multirow{-2}{*}{LCL} & L1 & L2 & L3 & L4 & L1 & L2 & L3 & L4 &L1 & L2 & L3 & L4 &L1 & L2 & L3 & L4 \\
\midrule
\multirow{-1}{*}{Xcube~\tablecite{ren2024xcube}} &  $>$10000  & 98.968 &   98.799 &  98.615 & 98.226 & 95.525 & 93.872 & 92.322 & 85.365 & 6.315 & 6.288 & 7.935 & 9.926 & 1.579 & 1.432 & 1.430 & 1.679\\
\midrule
\rowcolor{white} \multirow{-1}{*}{Xcube$^\dag$~\tablecite{ren2024xcube}} & $>$10000   & 99.393 &  99.794 &  99.824 & 99.079 & 96.753 &  95.535 & 93.422 & 87.365 & 4.015 & 4.142 & 5.740 & 7.627 & 1.543 & 1.408 & 1.259 & 1.639\\

\midrule
\midrule 
\multirow{-1}{*} {VecSet~\tablecite{shape2vecset}} &  512  & 94.768 & 88.890 & 80.126 & 59.347  & 77.545 & 67.929 & 55.516 & 34.619 & 27.380 & 42.075 & 100.975 & 159.151 & 2.939 & 3.056 & 3.470 & 6.034\\
\midrule
\rowcolor{white} \multirow{-1}{*}{Craftsman~\tablecite{li2024craftsman}} &  256  & 98.016 & 95.874 & 91.756 & 81.739  & 87.994 & 82.549 & 73.000 & 57.379 & 4.389 & 9.129 & 14.530 & 33.441 & 1.906 & 1.873 & 2.191 & 3.933\\
\midrule
 &  1280  & 99.964 & 99.925 & 99.678 & 97.890 & 96.561 & 95.975 & 91.618 & 83.124 & 2.236 & 2.506 & 4.444 & 6.432 & 1.448 & 1.215 & 1.205 & 1.828\\
\rowcolor{white} \multirow{-2}{*}{\textbf{Ours}} \multirow{-2}{*}{\makecell{\tiny{w/o DCA}\\ \tiny{w/o SES,DCA}}} & 1280   & 99.944 &  99.814 & 97.294 & 96.779 & 95.977 &  94.623 & 88.406 & 79.240 & 2.422 & 2.983 & 3.980 & 6.196 & 1.496 & 1.313 & 1.352 & 2.207\\   
\midrule
 &  256  & 99.507 & 98.986 & 96.669 & 89.577 & 93.272 & 90.466 & 82.386 & 68.669 & 3.356 & 5.202 & 10.276 & 24.527 & 1.555 & 1.410 & 1.618 & 3.035 \\
\rowcolor{white} \multirow{-2}{*}{\textbf{Ours}}\multirow{-2}{*}{~~~\makecell{\tiny{full}}} & 1280   & \textbf{99.988} &  \textbf{99.955} &  \textbf{99.880} & \textbf{99.170} & \textbf{97.038} &  \textbf{96.831} & \textbf{93.458} & \textbf{87.473} & \textbf{2.097} & \textbf{2.500} & \textbf{3.945} & \textbf{5.265} & \textbf{1.433} & \textbf{1.186} & \textbf{1.137} & \textbf{1.579} \\ 

\bottomrule
\end{tabular}

\begin{tikzpicture}[overlay, remember picture, shorten >=.5pt, shorten <=.5pt, transform canvas={yshift=.25\baselineskip}]

\end{tikzpicture}
\vspace{-5pt}
\caption{Quantitative comparison in Dora-bench. $^\dag$ indicates the fine-tuning model that uses the same training data as ours.}
\label{tab:quantitative_comparison_supp}
\vspace{-5pt}
\end{table*}
\begin{table*}[ht]
    \centering
    \begin{tabular}{|p{12cm}|} 
        \hline
        \texttt{Objaverse/000-002/0aecee43ac2749499a16ab4388a0baa2} \\
        \texttt{ABO/B07MF1TQYR} \\
        \texttt{Meta/meta\_ADT\_1\_0\_WireShelvingUnitBlackSmall\_1\_3d} \\
        \texttt{Objaverse/000-009/yC4xcavDg89GZgqjJtUs5KYbrgz} \\
        \texttt{ABO/B07V4FNHCD} \\
        \texttt{Objaverse/000-149/61ace9488e1c45718530e2d8f9da4a9d} \\
        \texttt{GSO/Sootheze\_Cold\_Therapy\_Elephant} \\
        \texttt{ABO/B07JYN8DBM} \\
        \texttt{ABO/B07N6Q9JB1} \\
        \texttt{ABO/B07XJB28C7} \\
        \texttt{Meta/meta\_DTC\_1\_0\_BasketPlasticRectangular\_3d} \\
        \texttt{Objaverse/000-008/8f0d1d4df2d64d1aa7c062868ca09535} \\
        \texttt{GSO/Rubbermaid\_Large\_Drainer} \\
        \texttt{Meta/meta\_DTC\_1\_0\_Pottery\_B0CJJ59SLH\_BlueHairFairy\_3d} \\
        \hline
    \end{tabular}
    \vspace{-3pt}
    \caption{Data sources for Figure \ref{fig:supplementary_qualitative_comparison} and \ref{fig:supplementary_qualitative_comparison1}}
    \vspace{-5pt}
    \label{tab:examples}
\end{table*}
This supplementary material provides additional details and results to complement our main paper. We first present implementation details (Appendix \ref{appendix:implementatnions}), followed by extensive comparisons of VAE performance (Appendix \ref{appendix:comparison_vae}) and 3D generation comparisons between our method and baselines (Appendix \ref{appendix:image3d}). We conclude with a discussion of the limitations and future work in Appendix \ref{appendix:limitation}.

\begin{appendix}
\section{More implementation details}
\label{appendix:implementatnions}

\noindent\textbf{Data Processing.} 
Our training data consists of approximately 400,000 3D meshes carefully filtered from Objaverse~\cite{objaverse}. Following CLAY~\cite{zhang2024clay}, we preprocess all meshes to ensure watertight geometry. The dataset is randomly split into training and test sets, where the test set is further utilized to construct our Dora-bench benchmark.

\noindent\textbf{Dora-bench Construction.} 
We introduce Dora-bench, a comprehensive benchmark designed to evaluate 3D reconstruction quality across different levels of geometric complexity. The benchmark integrates data from multiple sources: ABO \cite{collins2022abo}, GSO \cite{downs2022google}, Meta \cite{meta_dtc}, and Objaverse \cite{objaverse} test set. 
The benchmark categorizes models into four detail levels (Level 1 to Level 4), with approximately 800 samples per level. Due to the scarcity of highly detailed models in ABO, GSO, and Meta datasets, Level 4 samples are predominantly sourced from the Objaverse test set.

\noindent\textbf{Evaluation Metrics.}
We employ multiple complementary metrics to comprehensively evaluate reconstruction quality. To assess fine-grained geometric details, we compute the Sharp Normal Error (SNE) by rendering normal maps from 22 fixed, evenly spaced viewpoints around each object using nvdiffrast~\cite{Laine2020diffrast}. For quantitative evaluation of overall geometric accuracy, we utilize the Kaolin library~\cite{KaolinLibrary} to compute two additional metrics: F-score, which measures the coverage and completeness of the reconstructed shape, and Chamfer Distance (CD), which evaluates the bi-directional similarity between the reconstructed and ground truth point clouds.

\noindent\textbf{VAE Architecture and Training.} 
Our VAE architecture follows recent successful designs~\cite{li2024craftsman,zhao2023michelangelo}, with 8 self-attention layers in the encoder and 16 in the decoder.
For sharp edge sampling, we set the number of sampled points \( N_d = 32768 \), target sharp points \( N_{\text{desired}} = 16384 \) and angle threshold $\tau=30$. 
Following 3DShape2VecSet~\cite{shape2vecset}, we construct \( Q_{\text{space}}\) by combining two types of point sampling: points randomly sampled near the mesh surface and points uniformly sampled within the spatial range of [-1,1]. 

We adopt the multi-resolution training strategy proposed in CLAY~\cite{zhang2024clay}, where the latent code length (LCL) \( N_s \) is randomly selected between 256 and 1280 during training. This approach facilitates progressive training in the subsequent diffusion stage. The KL divergence weight is set to 0.001. We train our Dora-VAE on the Objaverse~\cite{objaverse} training set using 32 A100 GPUs with a batch size of 2048 for two days.

\noindent\textbf{Diffusion Model for Image-to-3D.} 
We apply our Dora-VAE to the downstream image-to-3D task. Specifically, we implement a conditional diffusion model based on the DiT architecture~\cite{Peebles2022DiT, chen2023pixartalpha}, similar to Direct3D~\cite{wu2024direct3d} and CLAY~\cite{zhang2024clay}. The model conditions on image features extracted by DINOv2~\cite{oquab2023dinov2} from single-view images rendered using BlenderProc~\cite{blenderproc}. Our diffusion model contains 0.39 billion parameters and is trained on 32 A100 GPUs for three days.

\section{More comparison of VAE}
\label{appendix:comparison_vae}
We present comprehensive quantitative and qualitative comparisons of our Dora-VAE against existing methods on the Dora-bench dataset. In addition to the baselines discussed in the main paper, we include 3DShape2VecSet~\cite{shape2vecset}, which was trained on ShapeNet~\cite{chang2015shapenet} rather than the larger Objaverse~\cite{objaverse} dataset.

\noindent\textbf{Quantitative Results.} We see in Table~\ref{tab:quantitative_comparison_supp}, 3DShape2VecSet~\cite{shape2vecset} consistently underperforms across all detail levels, primarily due to its limited training data affecting generalization capability. 

\begin{figure*}
\center
  \includegraphics[width=\textwidth]{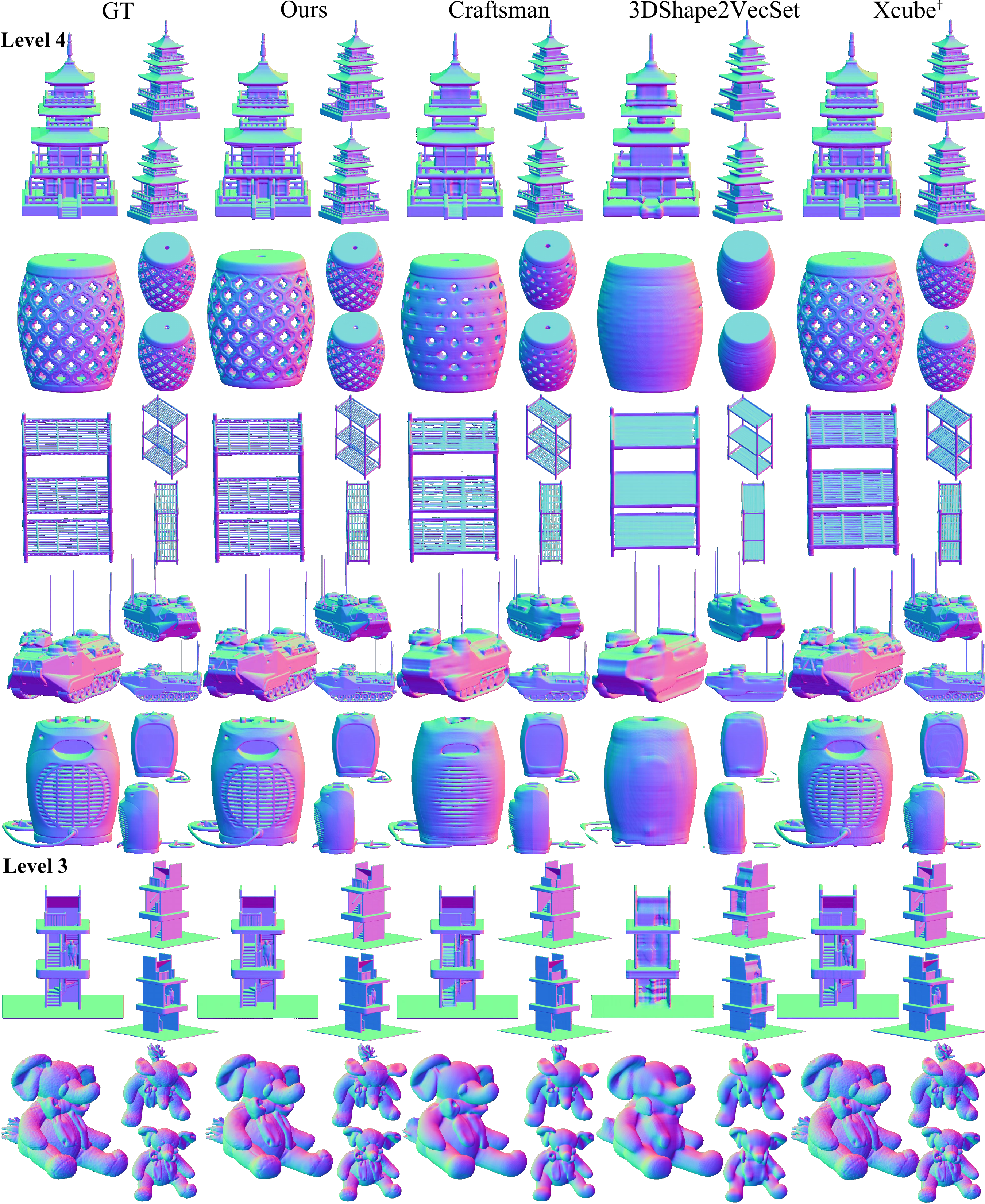}
  \caption{
    Qualitative comparison of the VAE reconstruction results. $^\dag$ indicates the fine-tuning model that uses the same training data as ours.
  }
  \label{fig:supplementary_qualitative_comparison}
\end{figure*}

\begin{figure*}
\center
  \includegraphics[width=\textwidth]{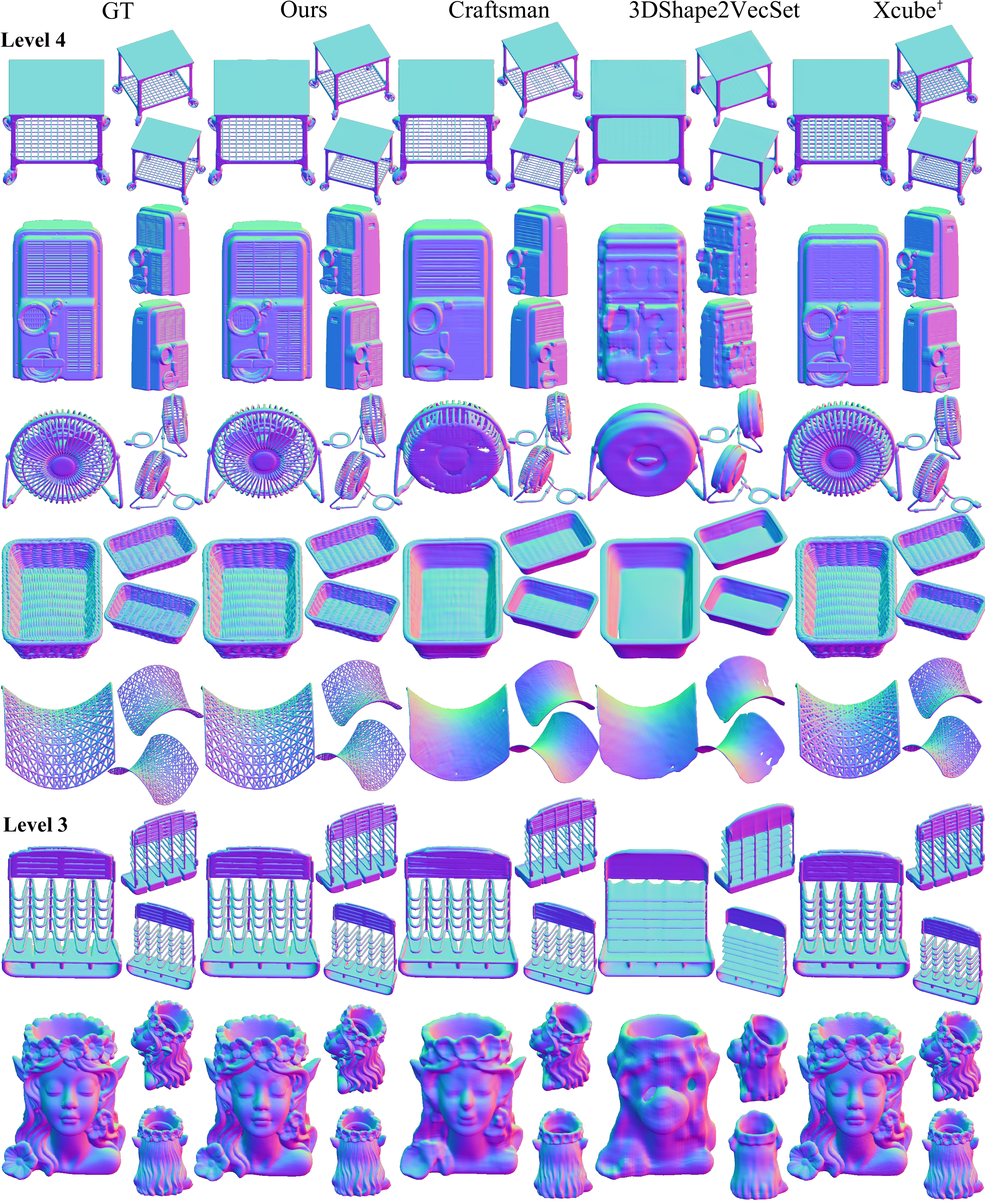}
  \caption{
    Qualitative comparison of the VAE reconstruction results. $^\dag$ indicates the fine-tuning model that uses the same training data as ours.
  }
  \label{fig:supplementary_qualitative_comparison1}
\end{figure*}

\noindent\textbf{Qualitative Evaluation.} 
Figures~\ref{fig:supplementary_qualitative_comparison} and \ref{fig:supplementary_qualitative_comparison1} present visual comparisons for Level 3 and 4 examples (specific data sources listed in Table~\ref{tab:examples}). For XCube~\cite{ren2024xcube}, we present only its fine-tuned version (XCube$^\dag$) as it slightly outperforms the original version. We see our Dora-VAE outperforms all other baselines. 
While XCube demonstrates rich visual details, we observe that its geometry sometimes deviates from the ground truth mesh. We attribute this to quantization errors introduced during mesh extraction using NKSR~\cite{huang2023neural}, which explains its lower performance in metrics like chamfer distance (CD) and SNE despite visually appealing results.

\section{Image-to-3D Generation Comparison}
\label{appendix:image3d}
We evaluate our Dora-VAE-based latent diffusion model against state-of-the-art methods for single-image 3D generation. Our comparison includes 1) LRM-based methods: MeshFormer~\cite{liu2024meshformer} and CRM~\cite{wang2024crm}, as well as 2) industry solution: Tripo v2.0~\cite{Tripo}. We use the official code and model provided by CRM \cite{wang2024crm} for inference and obtain the results of MeshFormer \cite{liu2024meshformer} and Tripo v2.0 \cite{Tripo} from their huggingface demo and product website. 
Note that CLAY~\cite{zhang2024clay} and Rodin Gen-1~\cite{Roding_gen1} are excluded due to implementation unavailability and usage limitations at submission time.

\begin{figure*}
\center
  \includegraphics[width=\textwidth]{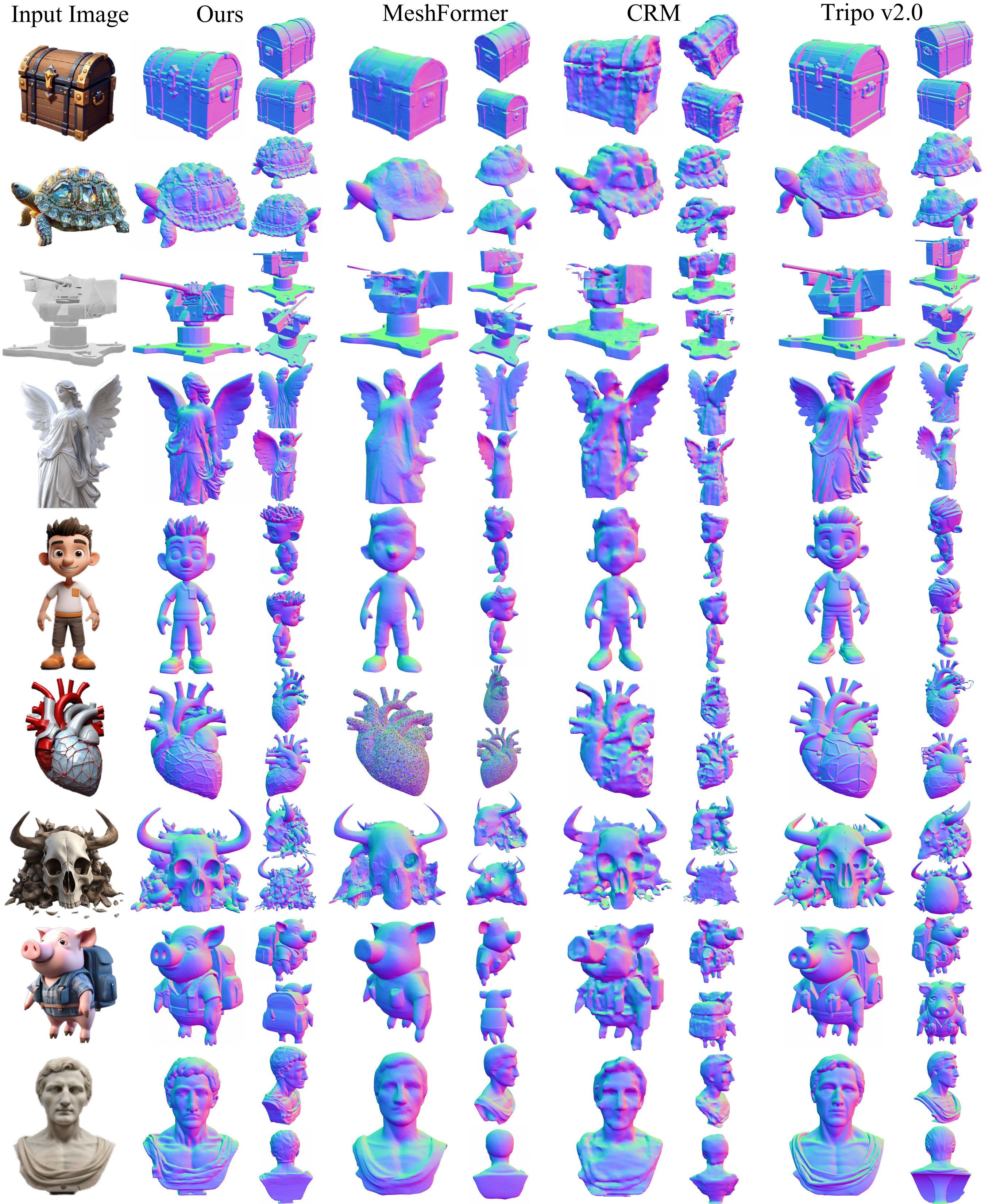}
  \caption{
    Qualitative comparison of the Image-to-3D results.
  }
  \label{fig:supplementary_diffusion_comp}
\end{figure*}

\begin{figure*}
\begin{minipage}[b]{1.\linewidth}
\center
  \includegraphics[width=1.\textwidth]{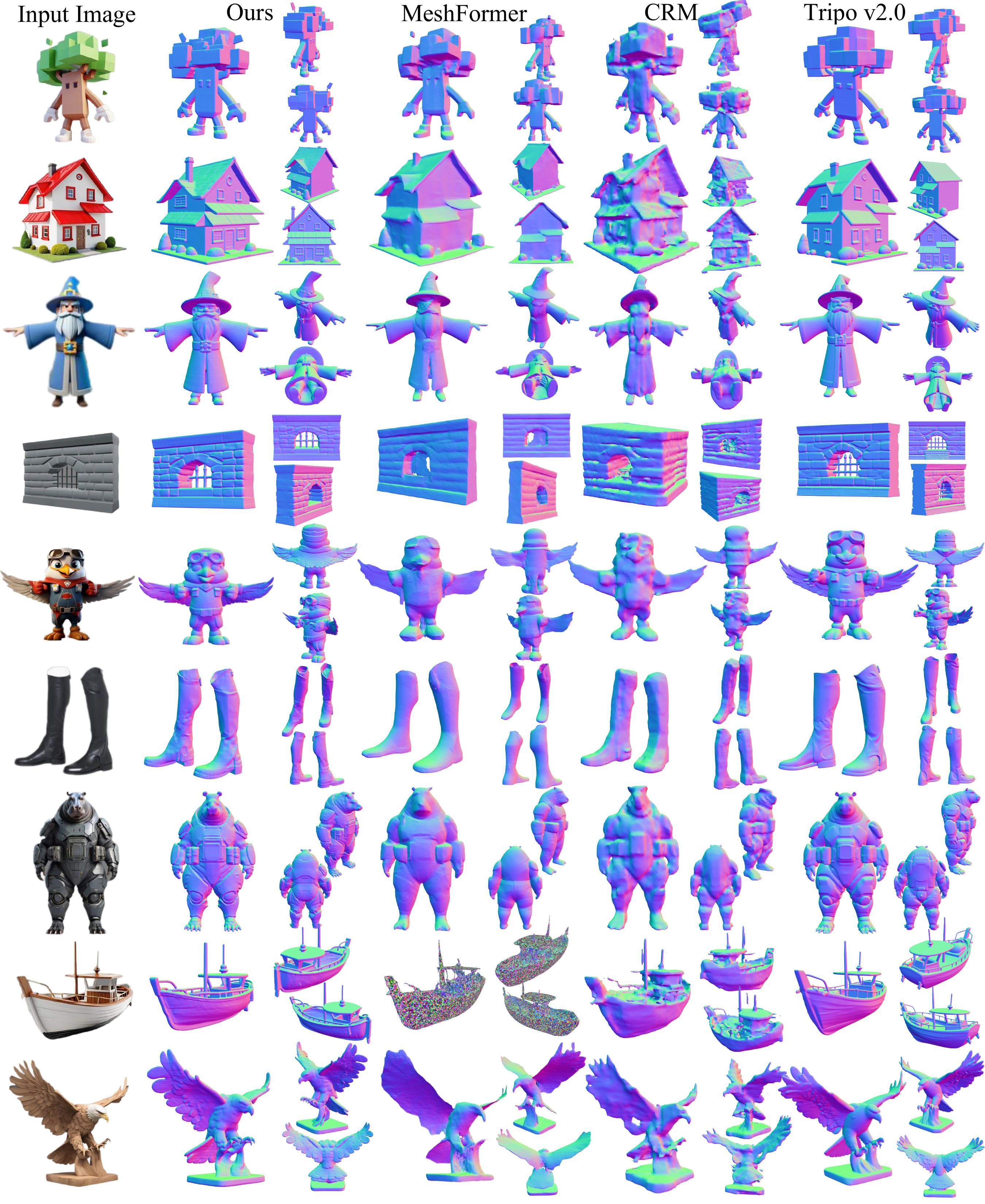}
  \caption{
    Qualitative comparison of the Image-to-3D results.
  }
  \label{fig:supplementary_diffusion1_comp}
  \vspace{-10cm}
\end{minipage}
\end{figure*}

As demonstrated in Figures~\ref{fig:supplementary_diffusion_comp} and~\ref{fig:supplementary_diffusion1_comp}, our method achieves superior results compared to LRM-based approaches in terms of both geometric detail and fidelity. The performance limitations of MeshFormer and CRM can be attributed to their lack of explicit geometric constraints, leading to unstable or lower-quality reconstructions.

Our method achieves comparable geometric quality to Tripo v2.0, a leading commercial solution, while using significantly more constrained resources. Specifically, we achieve these results with only three days of training on 32 A100 GPUs and approximately 400,000 training samples. This remarkable performance, achieved with limited computational resources and training data compared to commercial solutions, demonstrates the effectiveness of Dora-VAE in enhancing geometric detail and improving diffusion model performance.

\section{Limitations and Future Directions}
\label{appendix:limitation}
While Dora-VAE achieves state-of-the-art reconstruction quality with 1,280 latent code tokens, we identify several limitations and promising directions for future research.

\noindent\textbf{Current Limitations.}
The primary limitation of our approach lies in maintaining high-quality reconstructions when further reducing the number of latent tokens. This challenge becomes particularly evident when comparing with recent advances in 2D domain, such as Deep Compression Autoencoder (DC-AE)~\cite{chen2024dcae}, which has achieved remarkable compression rates while preserving reconstruction quality.

\noindent\textbf{Future Directions.}
We envision two main directions for future work:
1) \textit{Enhanced Compression Efficiency:}
We aim to explore novel techniques for increasing the compression rate of 3D VAEs while maintaining reconstruction quality. This research direction could potentially bridge the efficiency gap between 2D and 3D compression methods.
2) \textit{Advanced Diffusion Models:}
Building upon Dora-VAE's superior reconstruction capabilities, we plan to develop more powerful image-to-3D diffusion models. We believe that the improved reconstruction quality offered by Dora-VAE can directly boost the performance ceiling of diffusion models, enabling higher-quality generation results under the same training conditions.

\end{appendix}

\clearpage
{
    \small
    \bibliographystyle{ieeenat_fullname}
    \bibliography{bibliography}

\begin{thebibliography}{67}
\providecommand{\natexlab}[1]{#1}
\providecommand{\url}[1]{\texttt{#1}}
\expandafter\ifx\csname urlstyle\endcsname\relax
  \providecommand{\doi}[1]{doi: #1}\else
  \providecommand{\doi}{doi: \begingroup \urlstyle{rm}\Url}\fi

\bibitem[Rod(2024)]{Roding_gen1}
Rodin gen-1, 2024.
\newblock \url{https://hyperhuman.deemos.com/rodin/}.

\bibitem[Tri(2024)]{Tripo}
Tripo ai, 2024.
\newblock \url{https://www.tripo3d.ai/}.

\bibitem[met(2024)]{meta_dtc}
Digital twin catalog.
\newblock META, 2024.
\newblock \url{https://www.projectaria.com/datasets/dtc/}.

\bibitem[Cao et~al.(2024{\natexlab{a}})Cao, Cao, Han, Shan, and Wong]{cao2024dreamavatar}
Yukang Cao, Yan-Pei Cao, Kai Han, Ying Shan, and Kwan-Yee~K. Wong.
\newblock Dreamavatar: Text-and-shape guided 3d human avatar generation via diffusion models.
\newblock In \emph{Proceedings of the IEEE/CVF Conference on Computer Vision and Pattern Recognition}, pages 958--968, 2024{\natexlab{a}}.

\bibitem[Cao et~al.(2024{\natexlab{b}})Cao, Pan, Han, Wong, and Liu]{cao2024avatargo}
Yukang Cao, Liang Pan, Kai Han, Kwan-Yee~K Wong, and Ziwei Liu.
\newblock Avatargo: Zero-shot 4d human-object interaction generation and animation.
\newblock \emph{arXiv preprint arXiv:2410.07164}, 2024{\natexlab{b}}.

\bibitem[Chang et~al.(2015)Chang, Funkhouser, Guibas, Hanrahan, Huang, Li, Savarese, Savva, Song, Su, et~al.]{chang2015shapenet}
Angel~X Chang, Thomas Funkhouser, Leonidas Guibas, Pat Hanrahan, Qixing Huang, Zimo Li, Silvio Savarese, Manolis Savva, Shuran Song, Hao Su, et~al.
\newblock Shapenet: An information-rich 3d model repository.
\newblock \emph{arXiv preprint arXiv:1512.03012}, 2015.

\bibitem[Chen et~al.(2023{\natexlab{a}})Chen, Yu, Ge, Yao, Xie, Wu, Wang, Kwok, Luo, Lu, and Li]{chen2023pixartalpha}
Junsong Chen, Jincheng Yu, Chongjian Ge, Lewei Yao, Enze Xie, Yue Wu, Zhongdao Wang, James Kwok, Ping Luo, Huchuan Lu, and Zhenguo Li.
\newblock Pixart-$\alpha$: Fast training of diffusion transformer for photorealistic text-to-image synthesis, 2023{\natexlab{a}}.

\bibitem[Chen et~al.(2024{\natexlab{a}})Chen, Cai, Chen, Xie, Yang, Tang, Li, Lu, and Han]{chen2024dcae}
Junyu Chen, Han Cai, Junsong Chen, Enze Xie, Shang Yang, Haotian Tang, Muyang Li, Yao Lu, and Song Han.
\newblock Deep compression autoencoder for efficient high-resolution diffusion models.
\newblock \emph{arXiv preprint arXiv:2410.10733}, 2024{\natexlab{a}}.

\bibitem[Chen et~al.(2024{\natexlab{b}})Chen, Wu, Luo, Xie, Paul, Luo, Zhao, and Li]{chen2024pixartdelta}
Junsong Chen, Yue Wu, Simian Luo, Enze Xie, Sayak Paul, Ping Luo, Hang Zhao, and Zhenguo Li.
\newblock Pixart-$\delta$: Fast and controllable image generation with latent consistency models, 2024{\natexlab{b}}.

\bibitem[Chen et~al.(2023{\natexlab{b}})Chen, Chen, Jiao, and Jia]{fantasia3d}
Rui Chen, Yongwei Chen, Ningxin Jiao, and Kui Jia.
\newblock Fantasia3d: Disentangling geometry and appearance for high-quality text-to-3d content creation.
\newblock In \emph{Proceedings of the IEEE/CVF International Conference on Computer Vision (ICCV)}, 2023{\natexlab{b}}.

\bibitem[Chen et~al.(2024{\natexlab{c}})Chen, Shi, Huang, Tan, Komura, and Chen]{camdm}
Rui Chen, Mingyi Shi, Shaoli Huang, Ping Tan, Taku Komura, and Xuelin Chen.
\newblock Taming diffusion probabilistic models for character control.
\newblock In \emph{ACM SIGGRAPH 2024 Conference Papers}, New York, NY, USA, 2024{\natexlab{c}}. Association for Computing Machinery.

\bibitem[Chen et~al.(2016)Chen, Xu, Zhang, and Guestrin]{GradientCheckpoint}
Tianqi Chen, Bing Xu, Chiyuan Zhang, and Carlos Guestrin.
\newblock Training deep nets with sublinear memory cost.
\newblock \emph{CoRR}, abs/1604.06174, 2016.

\bibitem[Chen et~al.(2022)Chen, Chen, Lei, Zhang, and Jia]{chen2022tango}
Yongwei Chen, Rui Chen, Jiabao Lei, Yabin Zhang, and Kui Jia.
\newblock Tango: Text-driven photorealistic and robust 3d stylization via lighting decomposition.
\newblock \emph{Advances in Neural Information Processing Systems}, 35:\penalty0 30923--30936, 2022.

\bibitem[Collins et~al.(2022)Collins, Goel, Deng, Luthra, Xu, Gundogdu, Zhang, Vicente, Dideriksen, Arora, et~al.]{collins2022abo}
Jasmine Collins, Shubham Goel, Kenan Deng, Achleshwar Luthra, Leon Xu, Erhan Gundogdu, Xi Zhang, Tomas F~Yago Vicente, Thomas Dideriksen, Himanshu Arora, et~al.
\newblock Abo: Dataset and benchmarks for real-world 3d object understanding.
\newblock In \emph{Proceedings of the IEEE/CVF conference on computer vision and pattern recognition}, pages 21126--21136, 2022.

\bibitem[Dao(2024)]{FlashAttentionv2}
Tri Dao.
\newblock Flashattention-2: Faster attention with better parallelism and work partitioning.
\newblock In \emph{The Twelfth International Conference on Learning Representations, {ICLR} 2024, Vienna, Austria, May 7-11, 2024}. OpenReview.net, 2024.

\bibitem[Deitke et~al.(2022)Deitke, Schwenk, Salvador, Weihs, Michel, VanderBilt, Schmidt, Ehsani, Kembhavi, and Farhadi]{objaverse}
Matt Deitke, Dustin Schwenk, Jordi Salvador, Luca Weihs, Oscar Michel, Eli VanderBilt, Ludwig Schmidt, Kiana Ehsani, Aniruddha Kembhavi, and Ali Farhadi.
\newblock Objaverse: A universe of annotated 3d objects.
\newblock \emph{arXiv preprint arXiv:2212.08051}, 2022.

\bibitem[Denninger et~al.(2023)Denninger, Winkelbauer, Sundermeyer, Boerdijk, Knauer, Strobl, Humt, and Triebel]{blenderproc}
Maximilian Denninger, Dominik Winkelbauer, Martin Sundermeyer, Wout Boerdijk, Markus Knauer, Klaus~H. Strobl, Matthias Humt, and Rudolph Triebel.
\newblock Blenderproc2: A procedural pipeline for photorealistic rendering.
\newblock \emph{Journal of Open Source Software}, 8\penalty0 (82):\penalty0 4901, 2023.

\bibitem[Downs et~al.(2022)Downs, Francis, Koenig, Kinman, Hickman, Reymann, McHugh, and Vanhoucke]{downs2022google}
Laura Downs, Anthony Francis, Nate Koenig, Brandon Kinman, Ryan Hickman, Krista Reymann, Thomas~B McHugh, and Vincent Vanhoucke.
\newblock Google scanned objects: A high-quality dataset of 3d scanned household items.
\newblock In \emph{2022 International Conference on Robotics and Automation (ICRA)}, pages 2553--2560. IEEE, 2022.

\bibitem[Fuji~Tsang et~al.()Fuji~Tsang, Shugrina, Lafleche, Perel, Loop, Takikawa, Modi, Zook, Wang, Chen, Shen, Gao, Jatavallabhula, Smith, Rozantsev, Fidler, State, Gorski, Xiang, Li, Li, and Lebaredian]{KaolinLibrary}
Clement Fuji~Tsang, Maria Shugrina, Jean~Francois Lafleche, Or Perel, Charles Loop, Towaki Takikawa, Vismay Modi, Alexander Zook, Jiehan Wang, Wenzheng Chen, Tianchang Shen, Jun Gao, Krishna~Murthy Jatavallabhula, Edward Smith, Artem Rozantsev, Sanja Fidler, Gavriel State, Jason Gorski, Tommy Xiang, Jianing Li, Michael Li, and Rev Lebaredian.
\newblock Kaolin: A pytorch library for accelerating 3d deep learning research.

\bibitem[Groh et~al.(2018)Groh, Wieschollek, and Lensch]{IDIS}
Fabian Groh, Patrick Wieschollek, and Hendrik P.~A. Lensch.
\newblock Flex-convolution (million-scale point-cloud learning beyond grid-worlds).
\newblock In \emph{Asian Conference on Computer Vision (ACCV)}, 2018.

\bibitem[Ho et~al.(2020)Ho, Jain, and Abbeel]{ho2020denoising}
Jonathan Ho, Ajay Jain, and Pieter Abbeel.
\newblock Denoising diffusion probabilistic models.
\newblock \emph{arXiv preprint arxiv:2006.11239}, 2020.

\bibitem[Hong et~al.(2024{\natexlab{a}})Hong, Zhang, Gu, Bi, Zhou, Liu, Liu, Sunkavalli, Bui, and Tan]{LRM}
Yicong Hong, Kai Zhang, Jiuxiang Gu, Sai Bi, Yang Zhou, Difan Liu, Feng Liu, Kalyan Sunkavalli, Trung Bui, and Hao Tan.
\newblock {LRM:} large reconstruction model for single image to 3d.
\newblock In \emph{The Twelfth International Conference on Learning Representations, {ICLR} 2024, Vienna, Austria, May 7-11, 2024}. OpenReview.net, 2024{\natexlab{a}}.

\bibitem[Hong et~al.(2024{\natexlab{b}})Hong, Zhang, Gu, Bi, Zhou, Liu, Liu, Sunkavalli, Bui, and Tan]{hong2024lrmlargereconstructionmodel}
Yicong Hong, Kai Zhang, Jiuxiang Gu, Sai Bi, Yang Zhou, Difan Liu, Feng Liu, Kalyan Sunkavalli, Trung Bui, and Hao Tan.
\newblock Lrm: Large reconstruction model for single image to 3d, 2024{\natexlab{b}}.

\bibitem[Huang et~al.(2023)Huang, Gojcic, Atzmon, Litany, Fidler, and Williams]{huang2023neural}
Jiahui Huang, Zan Gojcic, Matan Atzmon, Or Litany, Sanja Fidler, and Francis Williams.
\newblock Neural kernel surface reconstruction.
\newblock In \emph{Proceedings of the IEEE/CVF Conference on Computer Vision and Pattern Recognition}, pages 4369--4379, 2023.

\bibitem[Kerbl et~al.(2023)Kerbl, Kopanas, Leimk{\"u}hler, and Drettakis]{kerbl20233d}
Bernhard Kerbl, Georgios Kopanas, Thomas Leimk{\"u}hler, and George Drettakis.
\newblock 3d gaussian splatting for real-time radiance field rendering.
\newblock \emph{ACM Trans. Graph.}, 42\penalty0 (4):\penalty0 139--1, 2023.

\bibitem[Knapitsch et~al.(2017)Knapitsch, Park, Zhou, and Koltun]{fscore}
Arno Knapitsch, Jaesik Park, Qian-Yi Zhou, and Vladlen Koltun.
\newblock Tanks and temples: benchmarking large-scale scene reconstruction.
\newblock \emph{ACM Trans. Graph.}, 36\penalty0 (4), 2017.

\bibitem[Laine et~al.(2020)Laine, Hellsten, Karras, Seol, Lehtinen, and Aila]{Laine2020diffrast}
Samuli Laine, Janne Hellsten, Tero Karras, Yeongho Seol, Jaakko Lehtinen, and Timo Aila.
\newblock Modular primitives for high-performance differentiable rendering.
\newblock \emph{ACM Transactions on Graphics}, 39\penalty0 (6), 2020.

\bibitem[Lan et~al.(2024)Lan, Hong, Yang, Zhou, Meng, Dai, Pan, and Loy]{lan2024ln3diff}
Yushi Lan, Fangzhou Hong, Shuai Yang, Shangchen Zhou, Xuyi Meng, Bo Dai, Xingang Pan, and Chen~Change Loy.
\newblock Ln3diff: Scalable latent neural fields diffusion for speedy 3d generation.
\newblock In \emph{ECCV}, 2024.

\bibitem[Li et~al.(2023)Li, Tan, Zhang, Xu, Luan, Xu, Hong, Sunkavalli, Shakhnarovich, and Bi]{li2023instant3d}
Jiahao Li, Hao Tan, Kai Zhang, Zexiang Xu, Fujun Luan, Yinghao Xu, Yicong Hong, Kalyan Sunkavalli, Greg Shakhnarovich, and Sai Bi.
\newblock Instant3d: Fast text-to-3d with sparse-view generation and large reconstruction model.
\newblock \emph{arXiv preprint arXiv:2311.06214}, 2023.

\bibitem[Li et~al.(2024{\natexlab{a}})Li, Chen, Chen, and Tan]{sweetdreamer}
Weiyu Li, Rui Chen, Xuelin Chen, and Ping Tan.
\newblock Sweetdreamer: Aligning geometric priors in 2d diffusion for consistent text-to-3d.
\newblock \emph{International Conference on Learning Representations (ICLR)}, 2024{\natexlab{a}}.

\bibitem[Li et~al.(2024{\natexlab{b}})Li, Liu, Chen, Liang, Chen, Tan, and Long]{li2024craftsman}
Weiyu Li, Jiarui Liu, Rui Chen, Yixun Liang, Xuelin Chen, Ping Tan, and Xiaoxiao Long.
\newblock Craftsman: High-fidelity mesh generation with 3d native generation and interactive geometry refiner.
\newblock \emph{arXiv preprint arXiv:2405.14979}, 2024{\natexlab{b}}.

\bibitem[Liang et~al.(2023)Liang, Yang, Lin, Li, Xu, and Chen]{EnVision2023luciddreamer}
Yixun Liang, Xin Yang, Jiantao Lin, Haodong Li, Xiaogang Xu, and Yingcong Chen.
\newblock Luciddreamer: Towards high-fidelity text-to-3d generation via interval score matching.
\newblock \emph{arXiv preprint arXiv:2311.11284}, 2023.

\bibitem[Lin et~al.(2023)Lin, Gao, Tang, Takikawa, Zeng, Huang, Kreis, Fidler, Liu, and Lin]{lin2023magic3d}
Chen-Hsuan Lin, Jun Gao, Luming Tang, Towaki Takikawa, Xiaohui Zeng, Xun Huang, Karsten Kreis, Sanja Fidler, Ming-Yu Liu, and Tsung-Yi Lin.
\newblock Magic3d: High-resolution text-to-3d content creation.
\newblock In \emph{IEEE Conference on Computer Vision and Pattern Recognition ({CVPR})}, 2023.

\bibitem[Liu et~al.(2023{\natexlab{a}})Liu, Wu, Wei, Rao, and Duan]{liu2023sherpa3d}
Fangfu Liu, Diankun Wu, Yi Wei, Yongming Rao, and Yueqi Duan.
\newblock Sherpa3d: Boosting high-fidelity text-to-3d generation via coarse 3d prior, 2023{\natexlab{a}}.

\bibitem[Liu et~al.(2024{\natexlab{a}})Liu, Xu, Jin, Chen, Varma~T, Xu, and Su]{liu2023one2345}
Minghua Liu, Chao Xu, Haian Jin, Linghao Chen, Mukund Varma~T, Zexiang Xu, and Hao Su.
\newblock One-2-3-45: Any single image to 3d mesh in 45 seconds without per-shape optimization.
\newblock 2024{\natexlab{a}}.

\bibitem[Liu et~al.(2024{\natexlab{b}})Liu, Zeng, Wei, Shi, Chen, Xu, Zhang, Wang, Zhang, Liu, Wu, and Su]{liu2024meshformer}
Minghua Liu, Chong Zeng, Xinyue Wei, Ruoxi Shi, Linghao Chen, Chao Xu, Mengqi Zhang, Zhaoning Wang, Xiaoshuai Zhang, Isabella Liu, Hongzhi Wu, and Hao Su.
\newblock Meshformer: High-quality mesh generation with 3d-guided reconstruction model.
\newblock \emph{arXiv preprint arXiv:2408.10198}, 2024{\natexlab{b}}.

\bibitem[Liu et~al.(2023{\natexlab{b}})Liu, Lin, Zeng, Long, Liu, Komura, and Wang]{liu2023syncdreamer}
Yuan Liu, Cheng Lin, Zijiao Zeng, Xiaoxiao Long, Lingjie Liu, Taku Komura, and Wenping Wang.
\newblock Syncdreamer: Generating multiview-consistent images from a single-view image.
\newblock \emph{arXiv preprint arXiv:2309.03453}, 2023{\natexlab{b}}.

\bibitem[Mildenhall et~al.(2020)Mildenhall, Srinivasan, Tancik, Barron, Ramamoorthi, and Ng]{mildenhall2020nerf}
Ben Mildenhall, Pratul~P. Srinivasan, Matthew Tancik, Jonathan~T. Barron, Ravi Ramamoorthi, and Ren Ng.
\newblock Nerf: Representing scenes as neural radiance fields for view synthesis.
\newblock In \emph{European Conference on Computer Vision (ECCV)}, 2020.

\bibitem[Moenning and Dodgson(2003)]{moenning2003fast}
Carsten Moenning and Neil~A Dodgson.
\newblock Fast marching farthest point sampling.
\newblock Technical report, University of Cambridge, Computer Laboratory, 2003.

\bibitem[Oquab et~al.(2023)Oquab, Darcet, Moutakanni, Vo, Szafraniec, Khalidov, Fernandez, Haziza, Massa, El-Nouby, Howes, Huang, Xu, Sharma, Li, Galuba, Rabbat, Assran, Ballas, Synnaeve, Misra, Jegou, Mairal, Labatut, Joulin, and Bojanowski]{oquab2023dinov2}
Maxime Oquab, Timothée Darcet, Theo Moutakanni, Huy~V. Vo, Marc Szafraniec, Vasil Khalidov, Pierre Fernandez, Daniel Haziza, Francisco Massa, Alaaeldin El-Nouby, Russell Howes, Po-Yao Huang, Hu Xu, Vasu Sharma, Shang-Wen Li, Wojciech Galuba, Mike Rabbat, Mido Assran, Nicolas Ballas, Gabriel Synnaeve, Ishan Misra, Herve Jegou, Julien Mairal, Patrick Labatut, Armand Joulin, and Piotr Bojanowski.
\newblock Dinov2: Learning robust visual features without supervision, 2023.

\bibitem[Peebles and Xie(2022)]{Peebles2022DiT}
William Peebles and Saining Xie.
\newblock Scalable diffusion models with transformers.
\newblock \emph{arXiv preprint arXiv:2212.09748}, 2022.

\bibitem[Peebles and Xie(2023)]{peebles2023scalable}
William Peebles and Saining Xie.
\newblock Scalable diffusion models with transformers.
\newblock In \emph{Proceedings of the IEEE/CVF International Conference on Computer Vision}, pages 4195--4205, 2023.

\bibitem[Poole et~al.(2022)Poole, Jain, Barron, and Mildenhall]{poole2022dreamfusion}
Ben Poole, Ajay Jain, Jonathan~T. Barron, and Ben Mildenhall.
\newblock Dreamfusion: Text-to-3d using 2d diffusion.
\newblock \emph{arXiv}, 2022.

\bibitem[Qiu et~al.(2024)Qiu, Chen, Gu, Zuo, Xu, Wu, Yuan, Dong, Bo, and Han]{RichDreamer}
Lingteng Qiu, Guanying Chen, Xiaodong Gu, Qi Zuo, Mutian Xu, Yushuang Wu, Weihao Yuan, Zilong Dong, Liefeng Bo, and Xiaoguang Han.
\newblock Richdreamer: {A} generalizable normal-depth diffusion model for detail richness in text-to-3d.
\newblock In \emph{{IEEE/CVF} Conference on Computer Vision and Pattern Recognition, {CVPR} 2024, Seattle, WA, USA, June 16-22, 2024}, pages 9914--9925. {IEEE}, 2024.

\bibitem[Ren et~al.(2024)Ren, Huang, Zeng, Museth, Fidler, and Williams]{ren2024xcube}
Xuanchi Ren, Jiahui Huang, Xiaohui Zeng, Ken Museth, Sanja Fidler, and Francis Williams.
\newblock Xcube: Large-scale 3d generative modeling using sparse voxel hierarchies.
\newblock In \emph{Proceedings of the IEEE/CVF Conference on Computer Vision and Pattern Recognition}, pages 4209--4219, 2024.

\bibitem[Rombach et~al.(2022)Rombach, Blattmann, Lorenz, Esser, and Ommer]{rombach2022high}
Robin Rombach, Andreas Blattmann, Dominik Lorenz, Patrick Esser, and Bj{\"o}rn Ommer.
\newblock High-resolution image synthesis with latent diffusion models.
\newblock In \emph{Conference on Computer Vision and Pattern Recognition (CVPR)}, pages 10684--10695, 2022.

\bibitem[Shen et~al.(2021)Shen, Gao, Yin, Liu, and Fidler]{shen2021deep}
Tianchang Shen, Jun Gao, Kangxue Yin, Ming-Yu Liu, and Sanja Fidler.
\newblock Deep marching tetrahedra: a hybrid representation for high-resolution 3d shape synthesis.
\newblock \emph{Advances in Neural Information Processing Systems}, 34:\penalty0 6087--6101, 2021.

\bibitem[Shi et~al.(2023)Shi, Wang, Ye, Mai, Li, and Yang]{shi2023MVDream}
Yichun Shi, Peng Wang, Jianglong Ye, Long Mai, Kejie Li, and Xiao Yang.
\newblock Mvdream: Multi-view diffusion for 3d generation.
\newblock \emph{arXiv:2308.16512}, 2023.

\bibitem[Tang et~al.(2024{\natexlab{a}})Tang, Chen, Chen, Wang, Zeng, and Liu]{tang2024lgm}
Jiaxiang Tang, Zhaoxi Chen, Xiaokang Chen, Tengfei Wang, Gang Zeng, and Ziwei Liu.
\newblock Lgm: Large multi-view gaussian model for high-resolution 3d content creation.
\newblock \emph{arXiv preprint arXiv:2402.05054}, 2024{\natexlab{a}}.

\bibitem[Tang et~al.(2024{\natexlab{b}})Tang, Ren, Zhou, Liu, and Zeng]{DreamGaussian}
Jiaxiang Tang, Jiawei Ren, Hang Zhou, Ziwei Liu, and Gang Zeng.
\newblock Dreamgaussian: Generative gaussian splatting for efficient 3d content creation.
\newblock In \emph{The Twelfth International Conference on Learning Representations, {ICLR} 2024, Vienna, Austria, May 7-11, 2024}. OpenReview.net, 2024{\natexlab{b}}.

\bibitem[Tochilkin et~al.(2024)Tochilkin, Pankratz, Liu, Huang, , Letts, Li, Liang, Laforte, Jampani, and Cao]{TripoSR2024}
Dmitry Tochilkin, David Pankratz, Zexiang Liu, Zixuan Huang, , Adam Letts, Yangguang Li, Ding Liang, Christian Laforte, Varun Jampani, and Yan-Pei Cao.
\newblock Triposr: Fast 3d object reconstruction from a single image.
\newblock \emph{arXiv preprint arXiv:2403.02151}, 2024.

\bibitem[Wang et~al.(2023{\natexlab{a}})Wang, Tan, Bi, Xu, Luan, Sunkavalli, Wang, Xu, and Zhang]{wang2023pf}
Peng Wang, Hao Tan, Sai Bi, Yinghao Xu, Fujun Luan, Kalyan Sunkavalli, Wenping Wang, Zexiang Xu, and Kai Zhang.
\newblock Pf-lrm: Pose-free large reconstruction model for joint pose and shape prediction.
\newblock \emph{arXiv preprint arXiv:2311.12024}, 2023{\natexlab{a}}.

\bibitem[Wang et~al.(2022)Wang, Zhang, Zhang, Gu, Bao, Baltrusaitis, Shen, Chen, Wen, Chen, and Guo]{wang2022rodingenerativemodelsculpting}
Tengfei Wang, Bo Zhang, Ting Zhang, Shuyang Gu, Jianmin Bao, Tadas Baltrusaitis, Jingjing Shen, Dong Chen, Fang Wen, Qifeng Chen, and Baining Guo.
\newblock Rodin: A generative model for sculpting 3d digital avatars using diffusion, 2022.

\bibitem[Wang et~al.(2023{\natexlab{b}})Wang, Lu, Wang, Bao, Li, Su, and Zhu]{wang2023prolificdreamer}
Zhengyi Wang, Cheng Lu, Yikai Wang, Fan Bao, Chongxuan Li, Hang Su, and Jun Zhu.
\newblock Prolificdreamer: High-fidelity and diverse text-to-3d generation with variational score distillation.
\newblock \emph{arXiv preprint arXiv:2305.16213}, 2023{\natexlab{b}}.

\bibitem[Wang et~al.(2024)Wang, Wang, Chen, Xiang, Chen, Yu, Li, Su, and Zhu]{wang2024crm}
Zhengyi Wang, Yikai Wang, Yifei Chen, Chendong Xiang, Shuo Chen, Dajiang Yu, Chongxuan Li, Hang Su, and Jun Zhu.
\newblock Crm: Single image to 3d textured mesh with convolutional reconstruction model.
\newblock \emph{arXiv preprint arXiv:2403.05034}, 2024.

\bibitem[Williams et~al.(2024)Williams, Huang, Swartz, Klar, Thakkar, Cong, Ren, Li, Fuji-Tsang, Fidler, Sifakis, and Museth]{williams2024fvdb}
Francis Williams, Jiahui Huang, Jonathan Swartz, Gergely Klar, Vijay Thakkar, Matthew Cong, Xuanchi Ren, Ruilong Li, Clement Fuji-Tsang, Sanja Fidler, Eftychios Sifakis, and Ken Museth.
\newblock fvdb: A deep-learning framework for sparse, large-scale, and high-performance spatial intelligence.
\newblock \emph{ACM Transactions on Graphics (TOG)}, 43\penalty0 (4):\penalty0 133:1--133:15, 2024.

\bibitem[Wu et~al.(2023)Wu, Zheng, Pfrommer, and Beyerer]{wu_2023_attention_edge}
Chengzhi Wu, Junwei Zheng, Julius Pfrommer, and J\"urgen Beyerer.
\newblock Attention-based point cloud edge sampling.
\newblock In \emph{Proceedings of the IEEE/CVF Conference on Computer Vision and Pattern Recognition (CVPR)}, 2023.

\bibitem[Wu et~al.(2024)Wu, Lin, Zhang, Zeng, Xu, Torr, Cao, and Yao]{wu2024direct3d}
Shuang Wu, Youtian Lin, Feihu Zhang, Yifei Zeng, Jingxi Xu, Philip Torr, Xun Cao, and Yao Yao.
\newblock Direct3d: Scalable image-to-3d generation via 3d latent diffusion transformer.
\newblock \emph{arXiv:2405.14832}, 2024.

\bibitem[Xiong et~al.(2024)Xiong, Wei, Zheng, Cao, Lian, and Wang]{xiong2024octfusion}
Bojun Xiong, Si-Tong Wei, Xin-Yang Zheng, Yan-Pei Cao, Zhouhui Lian, and Peng-Shuai Wang.
\newblock {OctFusion}: Octree-based diffusion models for 3d shape generation.
\newblock \emph{arXiv}, 2024.

\bibitem[Xu et~al.(2024)Xu, Cheng, Gao, Wang, Gao, and Shan]{xu2024instantmesh}
Jiale Xu, Weihao Cheng, Yiming Gao, Xintao Wang, Shenghua Gao, and Ying Shan.
\newblock Instantmesh: Efficient 3d mesh generation from a single image with sparse-view large reconstruction models.
\newblock \emph{arXiv preprint arXiv:2404.07191}, 2024.

\bibitem[Yuksel(2015)]{PoissionSampling}
Cem Yuksel.
\newblock Sample elimination for generating poisson disk sample sets.
\newblock \emph{The Eurographs Association \& John Wiley \& Sons, Ltd.}, 2015.

\bibitem[Zhang and Wonka(2024)]{zhang2024lagem}
Biao Zhang and Peter Wonka.
\newblock Lagem: A large geometry model for 3d representation learning and diffusion.
\newblock \emph{arXiv preprint arXiv:2410.01295}, 2024.

\bibitem[Zhang et~al.(2023{\natexlab{a}})Zhang, Tang, Nie\ss{}ner, and Wonka]{shape2vecset}
Biao Zhang, Jiapeng Tang, Matthias Nie\ss{}ner, and Peter Wonka.
\newblock 3dshape2vecset: A 3d shape representation for neural fields and generative diffusion models.
\newblock \emph{ACM Transactions on Graphics (SIGGRAPH)}, 42\penalty0 (4), 2023{\natexlab{a}}.

\bibitem[Zhang et~al.(2023{\natexlab{b}})Zhang, Rao, and Agrawala]{zhang2023adding}
Lvmin Zhang, Anyi Rao, and Maneesh Agrawala.
\newblock Adding conditional control to text-to-image diffusion models, 2023{\natexlab{b}}.

\bibitem[Zhang et~al.(2024)Zhang, Wang, Zhang, Qiu, Pang, Jiang, Yang, Xu, and Yu]{zhang2024clay}
Longwen Zhang, Ziyu Wang, Qixuan Zhang, Qiwei Qiu, Anqi Pang, Haoran Jiang, Wei Yang, Lan Xu, and Jingyi Yu.
\newblock Clay: A controllable large-scale generative model for creating high-quality 3d assets.
\newblock \emph{ACM Transactions on Graphics (TOG)}, 43\penalty0 (4):\penalty0 1--20, 2024.

\bibitem[Zhang et~al.(2023{\natexlab{c}})Zhang, Yin, Chen, Lin, Li, Hou, and Cheng]{zhang2023temo}
Xuying Zhang, Bo-Wen Yin, Yuming Chen, Zheng Lin, Yunheng Li, Qibin Hou, and Ming-Ming Cheng.
\newblock Temo: Towards text-driven 3d stylization for multi-object meshes.
\newblock \emph{arXiv preprint arXiv:2312.04248}, 2023{\natexlab{c}}.

\bibitem[Zhao et~al.(2023)Zhao, Liu, Chen, Zeng, Wang, Cheng, FU, Chen, YU, and Gao]{zhao2023michelangelo}
Zibo Zhao, Wen Liu, Xin Chen, Xianfang Zeng, Rui Wang, Pei Cheng, BIN FU, Tao Chen, Gang YU, and Shenghua Gao.
\newblock Michelangelo: Conditional 3d shape generation based on shape-image-text aligned latent representation.
\newblock In \emph{Advances in Neural Information Processing Systems (NeurIPS)}, 2023.

\end{thebibliography}
}

\end{document}